\pdfoutput=1

\documentclass[11pt]{article}

\usepackage[]{EMNLP2022}

\usepackage{times}
\usepackage{latexsym}
\usepackage{booktabs}
\usepackage{soul}
\usepackage{xcolor}
\usepackage{subcaption}
\usepackage{url}
\usepackage{graphicx}
\usepackage{multirow}
\usepackage{amssymb}
\usepackage{pifont}

\usepackage[T1]{fontenc}

\usepackage[utf8]{inputenc}

\usepackage{microtype}

\usepackage{siunitx}
\newcommand{\nop}[1]{}
\newcommand{\hlc}[2][yellow]{{%
    \colorlet{foo}{#1}%
    \sethlcolor{foo}\hl{#2}}%
}
\newcommand{\cmark}{\ding{51}}%
\newcommand{\xmark}{\ding{55}}%

\title{Thinking about GPT-3 In-Context Learning for Biomedical IE?\\ Think Again}

\author{Bernal Jim\'{e}nez Guti\'{e}rrez$^1$, Nikolas McNeal$^1$, Clay Washington$^1$,\\
\textbf{You Chen$^2$, Lang Li$^1$, Huan Sun$^1$, Yu Su$^1$}
\\$^1$The Ohio State University, $^2$Vanderbilt University \\
\small{\texttt{\{jimenezgutierrez.1,mcneal.121,washington.534,}}\small{\texttt{sun.397,su.809\}@osu.edu}}\\ \small{\texttt{lang.li@osumc.edu, you.chen@vumc.org}}
}

\begin{document}
\maketitle
\begin{abstract}
Large pre-trained language models (PLMs) such as GPT-3 have shown strong in-context learning capabilities, which are highly appealing for domains such as biomedicine that feature high and diverse demands of language technologies but also high data annotation costs. 
In this paper, we present the first systematic and comprehensive study to compare the few-shot performance of GPT-3 in-context learning with fine-tuning smaller (i.e., BERT-sized) PLMs on two representative biomedical information extraction (IE) tasks: named entity recognition and relation extraction. We follow the true few-shot setting~\cite{Perez2021TrueFL} to avoid overestimating models' few-shot performance by model selection over a large validation set. 
We also optimize GPT-3's performance with known techniques such as contextual calibration and dynamic in-context example retrieval. 
However, our results show that GPT-3 still significantly underperforms compared to simply fine-tuning a smaller PLM. 
In addition, GPT-3 in-context learning also yields smaller gains in accuracy when more training data becomes available.
More in-depth analyses further reveal issues of in-context learning that may be detrimental to IE tasks in general.
Given the high cost of experimenting with GPT-3, we hope our study provides helpful guidance for biomedical researchers and practitioners towards more practical solutions such as fine-tuning small PLMs before better in-context learning is available for biomedical IE.\footnote{Our source code is available at \url{https://github.com/dki-lab/few-shot-bioIE}.}

\end{abstract}

\section{Introduction}

\begin{figure}[!h]
 \centering
 \includegraphics[width=\columnwidth]{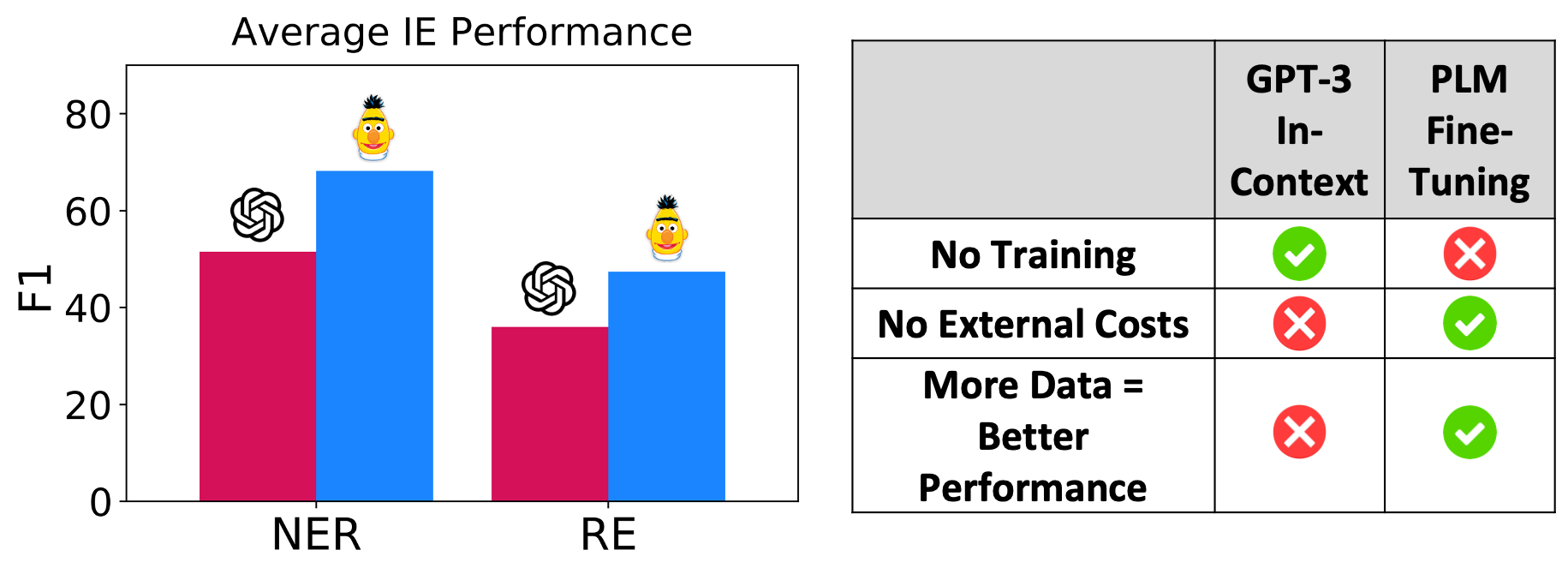}
 \vspace{-15pt}
\caption{Main findings: (\textit{Left}) fine-tuning BERT-sized PLMs substantially outperforms GPT-3 in-context learning under true few-shot setting. (\textit{Right}) Feature comparison for consideration of practical applications. }
        \label{fig:intro}
\vspace{-15pt}
\end{figure}

Given the overwhelming pace of biomedical research and clinical text production, transforming large amounts of biomedical text into structured information has become increasingly important for researchers and practitioners alike. In recent years, pre-trained language models (PLMs), both general-domain and biomedicine-specific ones, have remarkably boosted performance on biomedical information extraction (IE) tasks~\cite{Lee2020BioBERTAP, Peng2019TransferLI, Gu2022DomainSpecificLM, Alsentzer2019PubliclyAC, Beltagy2019SciBERTAP}.

The latest round of PLMs such as GPT-3~\cite{Brown2020LanguageMA}, Megatron-Turing NLG~\cite{megatron}, the Switch Transformer~\cite{switch}, among others, feature hundreds of billions of parameters and have achieved impressive performance in many NLP tasks using \textit{in-context learning}---a new few-shot learning paradigm first introduced by \citet{Brown2020LanguageMA}. In-context learning allows PLMs to use their natural language generation capabilities to solve any task almost like how humans would---by completing a piece of text or \textit{prompt}. This paradigm allows large PLMs to solve various NLP problems without updating their parameters, potentially resulting in massive savings in both data annotation and engineering overhead compared with standard model training. Even more impressively, GPT-3 in-context learning yields competitive performance against fully supervised baselines in many NLP tasks by adding only a handful of demonstrative examples in the prompt \cite{Brown2020LanguageMA}.

The variety of potential biomedical information extraction applications, the high cost of biomedical annotations, and the complexity of model training make in-context learning particularly appealing for biomedical applications. To investigate its practicality, we present the first systematic and comprehensive comparative study of GPT-3 in-context learning and BERT-sized \cite{Devlin2019BERTPO} PLM fine-tuning in the few-shot setting on named entity recognition (NER) and relation extraction (RE), two representative and highly valued biomedical IE tasks. For consistency and comprehensiveness, we use all the biomedical NER and RE tasks compiled in the BLURB benchmark \cite{Gu2022DomainSpecificLM}. We operate under the true few-shot setting introduced by \citet{Perez2021TrueFL} to avoid overestimating models' few-shot performance via model selection over a large validation set.

We optimize GPT-3's in-context learning performance for biomedical information extraction by leveraging multiple recent techniques. Firstly, inspired by studies that show the importance of optimal prompt selection \cite{Perez2021TrueFL, schick-schutze-2021-just, Gao2021MakingPL}, we formulate a prompt structure which allows us to construct prompt designs and select optimal ones systematically. Secondly, similar to \citet{liu-etal-2022-makes}, we introduce a k-nearest neighbor (kNN) module for in-context example retrieval. Finally, for NER, we also use logit biases to ensure that the generated tokens are from the input sentence; for RE, we use contextual calibration \cite{Zhao2021CalibrateBU} to reduce contextual bias.

Even when equipped with these latest techniques, which indeed improve GPT-3's performance as shown in ablation studies, we find that fine-tuning BERT-sized PLMs substantially outperforms GPT-3 in-context learning across all biomedical information extraction datasets when using the same small training set (e.g., \num{100} labeled examples). We also find that fine-tuning small PLMs yields a more reliable return in terms of data annotation: as training data size increases, fine-tuning performance steadily improves while in-context learning performance lags behind. In-depth analyses further reveal that in-context learning struggles with the \textit{\texttt{null} class}, e.g., sentences that contain no named entity (for NER) or entity pairs that hold none of the target relations (for RE), which is likely detrimental to IE tasks in general. In summary, our findings suggest that fine-tuning PLMs is still a more cost-effective option than GPT-3 in-context learning for biomedical IE tasks, at least before qualitatively better methods for in-context learning are discovered.

\begin{figure*}[!h]
     \centering
     \begin{subfigure}[b]{0.33\textwidth}
         \centering
         \includegraphics[width=\textwidth]{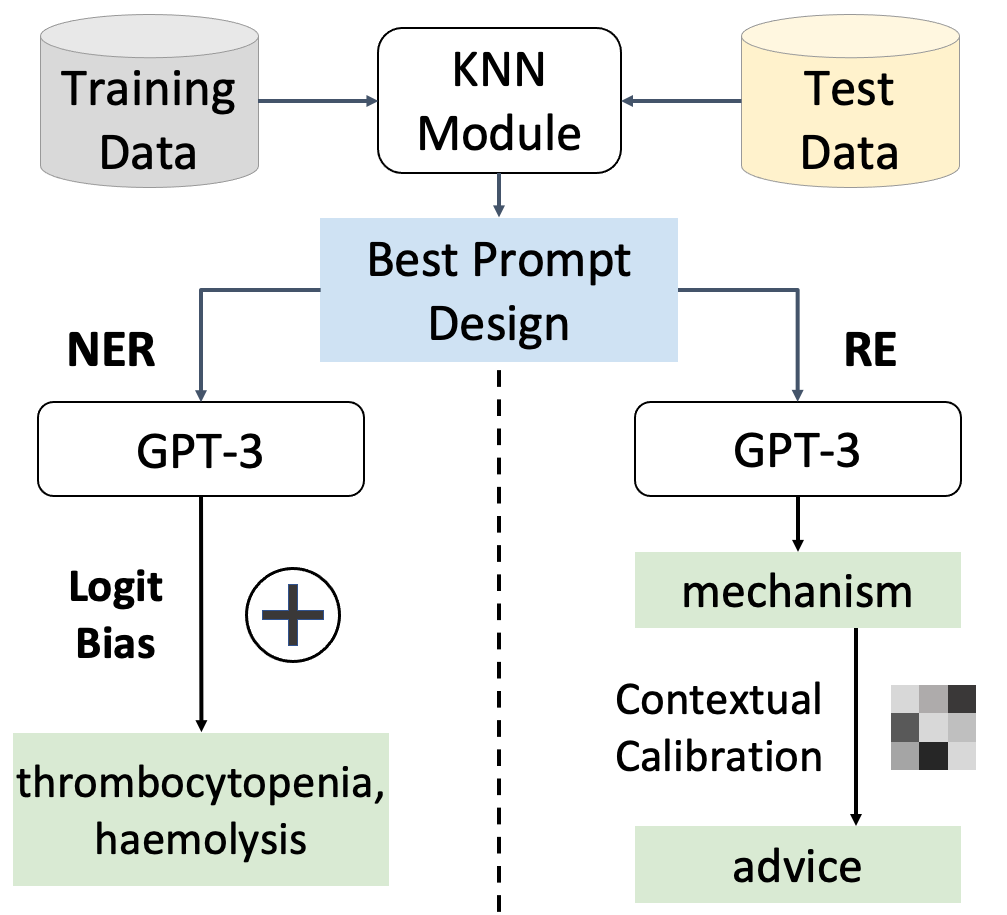}
         \label{fig:overall_system}
     \end{subfigure}
     \begin{subfigure}[b]{0.281\textwidth}
         \centering
         \includegraphics[width=\textwidth]{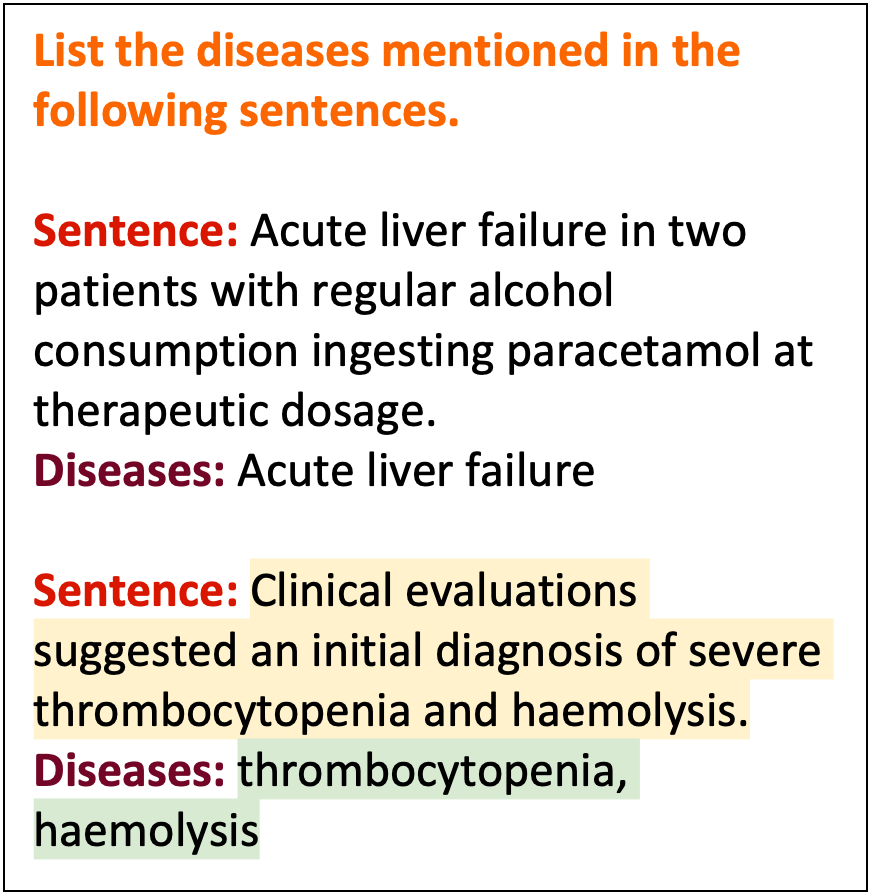}
         \label{fig:ner_prompt}
     \end{subfigure}
     \begin{subfigure}[b]{0.367\textwidth}
         \centering
         \includegraphics[width=\textwidth]{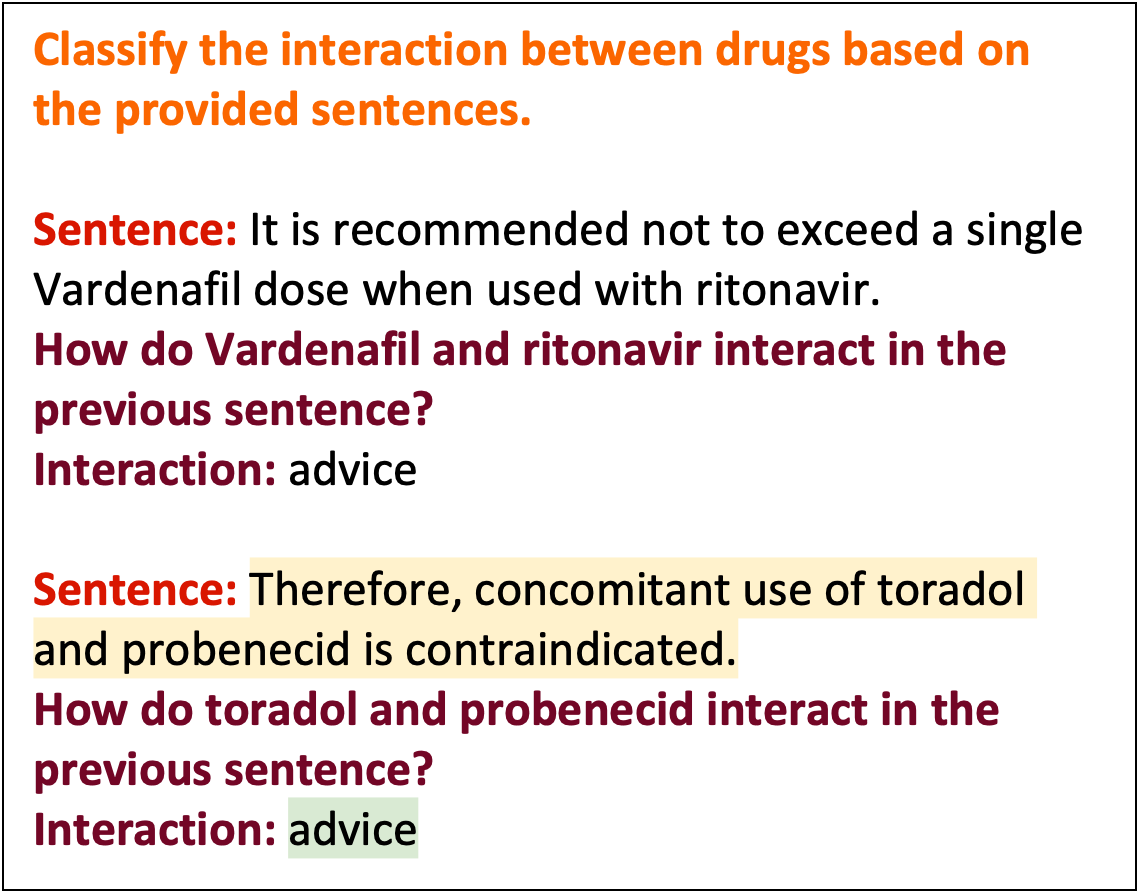}
         \label{fig:re_prompt}
     \end{subfigure}
     \vspace{-15pt}
        \caption{Overall architecture for GPT-3 in-context learning for both NER and RE (left). One-shot learning example prompt for NER (middle) and RE (right). Different colors indicate different prompt design elements: \textcolor{red!60!yellow}{orange} for overall task instructions, \textcolor{red!85!black}{red} for sentence introduction and \textcolor{purple!60!black}{purple} for the retrieval message portion. The current input sentence and the completion by GPT-3 are highlighted.}
        \label{fig:overall_system_and_examples}
\vspace{-12pt}
\end{figure*}

\section{Approach}

In this section, we describe the two paradigms we explored under the true few-shot setting for NER and RE: BERT-sized PLM fine-tuning and GPT-3 in-context learning.

\subsection{Tasks}

We use named entity recognition (NER) and relation extraction (RE) as two representative and highly valued tasks to comprehensively evaluate the potential of GPT-3 in-context learning in biomedical IE.

\subsection{True Few-Shot Setting}\label{sec:true-few-shot}

Recent work has questioned the performance of few-shot learning in very large PLMs like GPT-3 as well as small PLM fine-tuning, arguing that large validation sets have played a strong biasing role in model and prompt selection \cite{Perez2021TrueFL}. To avoid overestimating the few-shot learning performance of PLMs, we follow their proposed true few-shot setting. In this setting, all model selection decisions are made systematically on the few-shot training set rather than on a large validation set. For our main experiments, we use cross-validation on \num{100} training examples to choose the prompt structure, the number of few-shot examples per prompt and the fine-tuning hyperparameters.

\subsection{BERT-Sized PLM Fine-Tuning}

We follow the standard PLM fine-tuning process for NER and RE used in \citet{Gu2022DomainSpecificLM}. We use 5-fold cross-validation on the 100 example training set mentioned above to select the best performing values for learning rate, batch size, warm-up ratio, weight decay, and stopping checkpoint for all of our fine-tuning experiments. The hyperparameter values we select from are specified in Appendix~\ref{sec:appendix-hps}. 

\textbf{Named Entity Recognition.} For NER, we use the BIO tag token classification formulation and fine-tune a separate model for each entity type.

\textbf{Relation Extraction.} For RE, we mask the object and subject entities in the input sentence and use the \texttt{[CLS]} token to classify the relation between them.

\subsection{GPT-3 In-Context Learning}

In this section, we first describe how we reformulate the NER and RE tasks for in-context learning. We then provide thorough descriptions of our prompt design and in-context example retrieval approaches as well as other recent techniques we use to improve GPT-3's in-context learning performance for biomedical IE.

\subsubsection{Task Linearization}

As shown in the examples in Figure \ref{fig:overall_system_and_examples}, in order to use in-context learning, we must first reformulate NER and RE as language generation tasks. 

For NER, we extract all entity spans from the original sentence and combined them using a separator (entities are only added once), as was done in previous work \cite{Raval2021ExploringAU}. GPT-3 will then be expected to generate a list of entities joined by the chosen separator when conditioned on the current input and its context, as shown in Figure \ref{fig:overall_system_and_examples} (middle). 

For relation extraction, we draw inspiration from \citet{huguet-cabot-navigli-2021-rebel-relation} and transform every example into a prompt as shown in Figure \ref{fig:overall_system_and_examples} (right). For all our prompt templates shown in Appendix~\ref{sec:appendix-prompts}, we add the subject and object entities, in their verbatim lexical form in the original sentence, to the prompt.

\subsubsection{Prompt Design}\label{sec:prompt_design}

Given the importance of prompt selection in obtaining strong performance from GPT-3 in-context learning \cite{Perez2021TrueFL, schick-schutze-2022-true, schick-schutze-2021-just, Gao2021MakingPL}, we provide a systematic and task-agnostic process for constructing GPT-3 prompts. As shown in the examples in Figure \ref{fig:overall_system_and_examples}, we identify three main parts of a prompt: overall task instructions, a sentence introduction and a retrieval message. The overall task instruction provides broad instructions for the task as concisely as possible. The sentence introduction describes the input text (i.e., scientific article excerpt, tweet, sentence, etc.). Finally, the retrieval message directly precedes the expected completion and is meant to reiterate what is needed for the task. For relation extraction, similar to \citet{Schick2020AutomaticallyIW}, we also define a label verbalizer which maps relation categories to plausible natural language phrases to facilitate generation. 

For each task, we manually create a set of alternatives for each prompt section and select their best combination. We use leave-one-out cross-validation (LOOCV) to choose the best combination of the prompt alternatives as well as the number of in-context examples included in the prompt. To keep costs reasonable, we compare \num{8} prompt alternatives for each dataset. A list of all the options for each dataset can be found in Appendix~\ref{sec:appendix-prompts}.

\subsubsection{Logit Biases}

In order to prevent GPT-3 from generating tokens that are not in the original sentence, we use the \textit{logit bias} option from the OpenAI Completion API.\footnote{\url{https://beta.openai.com/docs/api-reference/completions}} This option allows us to add a fixed value to the final probability of a specified set of tokens, restricting the possible tokens that GPT-3 can generate. Specifically, we add a value of \num{10} to all tokens present in the original sentence, our chosen separator and the newline token (used to designate the end of the entity list). Additionally, any predicted entities that do not match any span in the original sentence are discarded during post-processing.

\subsubsection{Contextual Calibration}

During preliminary studies, we found that each set of few-shot in-context examples biased GPT-3 towards certain labels regardless of the test input. Previous work \cite{Zhao2021CalibrateBU} proposes to address these biases by calibrating the output using a linear transformation which equalizes all label probabilities generated by GPT-3 when conditioned on a null prompt (a version of the original prompt in which the test input is replaced by a null value such as ``N/A''). This linear transformation is then used to update the output probabilities of the true few-shot prompt, thereby removing the context induced biases. We adopt this approach for RE and create the null prompt by replacing the original sentence as well as the subject and object entities in the retrieval message with ``N/A''.

\subsubsection{Retrieval Module}

Several studies \cite{liu-etal-2022-makes, rubin-etal-2022-learning, shin-etal-2021-constrained} suggest that choosing few-shot in-context examples for each test example dynamically instead of using a fixed set of in-context examples yields strong improvements for GPT-3 in-context learning. Following \citet{liu-etal-2022-makes}, we use a k-nearest neighbor (kNN) retrieval module to select the most similar examples in our training set as the few-shot in-context prompt for each test example. We opt for RoBERTa-large as the encoder for our kNN retrieval module after preliminary experiments showing its advantages over other alternatives including biomedical PLMs \cite{Lee2020BioBERTAP,Gu2022DomainSpecificLM}, sentence-transformer models \cite{Reimers2019SentenceBERTSE} and a BM25 baseline \cite{bm25}.

\begin{table}[t]
\small
\centering
\resizebox{\columnwidth}{!}{%
\begin{tabular}{@{}lccccc@{}}
\toprule
   & 
  \textbf{Task} &
  \textbf{Train} &
  \textbf{Dev} &
  \textbf{Test} &
  \textbf{Eval.\ Metric}
   \\
  \midrule
\textbf{BC5CDR-disease} & NER & \num{4182} & \num{4244} & \num{4424} & F1 entity-level\\
\textbf{BC5CDR-chem} & NER & \num{5203} & \num{5347} & \num{5385} & F1 entity-level\\
\textbf{NCBI-disease} & NER &  \num{5134} & \num{787} & \num{960}  & F1 entity-level\\
\textbf{JNLPBA} & NER &  \num{46750} & \num{4551} & \num{8662} & F1 entity-level\\
\textbf{BC2GM} & NER & \num{15197} & \num{3061} & \num{6325} & F1 entity-level\\ 
\midrule
\midrule
\textbf{DDI} & RE & \num{25296} & \num{2496} & \num{5716} & Micro F1\\ 
\textbf{ChemProt} & RE & \num{18035} & \num{11268} & \num{15745} & Micro F1 \\
\textbf{GAD} & RE & \num{  4261  } & \num{  535  } & \num{  534  } & Micro F1\\
\bottomrule
\end{tabular}%
}
\caption{Dataset statistics.}

\label{tab:dataset_stats}
\vspace{-12pt}
\end{table}

\section{Experiments}

\subsection{Datasets}

We use all NER and RE datasets exactly as they are used in the BLURB benchmark  \citep{Gu2022DomainSpecificLM} to evaluate biomedical IE. Table \ref{tab:dataset_stats} lists the datasets and their statistics. For processing and train/dev/test splits,  we refer the interested reader to Section 2.3 of  \citet{Gu2022DomainSpecificLM}.

\subsubsection{Named Entity Recognition}

\noindent \textbf{BC5CDR.}
The BioCreative V Chemical-Disease Relation corpus \cite{jiaoli} contains PubMed abstracts with both disease and chemical annotations; we evaluate models on each entity type separately following previous work \cite{Gu2022DomainSpecificLM}.\\
\noindent \textbf{NCBI-disease.}
The Natural Center for Biotechnology Information Disease corpus \cite{ncbi} contains disease name and concept annotations for \num{793} PubMed abstracts. \\
\noindent \textbf{JNLPBA.}
The Joint Workshop on Natural Language Processing in Biomedicine and its Applications dataset \cite{jnlpba} contains \num{2000} abstracts from MEDLINE selected and annotated by hand for gene related entities. \\
\noindent \textbf{BC2GM.}
The Biocreative II Gene Mention corpus \cite{bc2gm} contains \num{17500} sentences from PubMed abstracts labeled for gene entities.

\begin{table*}[!ht]
\small
\centering
\resizebox{\textwidth}{!}{%
\begin{tabular}{@{}lcccc@{}}
\toprule
 &
  \textbf{PubMedBERT-base} &
  \textbf{BioBERT-large} &
  \textbf{RoBERTa-large} &
  \textbf{\begin{tabular}[c]{@{}c@{}}GPT-3 In-Context\end{tabular}} \\\cmidrule(l){2-5}
   & Precision / Recall / F1 & Precision / Recall / F1 & Precision / Recall / F1 & Precision / Recall / F1  \\\midrule
\textbf{BC5CDR-disease} & $ 67.4_{ 3.7}$/ $ 67.5_{ 1.2}$/ $ 67.4_{ 2.4}$ & $ 62.9_{ 5.0}$/ $ 69.0_{ 3.0}$/ $ 65.8_{ 4.1}$ & $ 66.9_{ 1.7}$/ $ 68.7_{ 4.7}$/ \textbf{67.7}$ _{ 1.8}$ & $ 57.9_{ 2.3}$/ $ 35.0_{ 2.9}$/ $ 43.6_{ 2.2}$\\
\textbf{BC5CDR-chem} & $ 86.1_{ 1.9}$/ $ 88.6_{ 4.8}$/ \textbf{87.3}$ _{ 1.3}$ & $ 84.8_{ 2.6}$/ $ 87.3_{ 3.3}$/ $ 86.0_{ 1.1}$ & $ 82.1_{ 1.8}$/ $ 87.3_{ 1.0}$/ $ 84.6_{ 1.3}$ & $ 74.7_{ 2.5}$/ $ 71.4_{ 2.2}$/ $ 73.0_{ 0.3}$\\
\textbf{NCBI-disease} & $ 68.5_{ 4.7}$/ $ 67.6_{ 2.4}$/ \textbf{68.0}$ _{ 2.9}$ & $ 59.6_{ 10.6}$/ $ 67.0_{ 6.1}$/ $ 63.0_{ 8.7}$ & $ 64.3_{ 3.7}$/ $ 68.7_{ 6.7}$/ $ 66.4_{ 5.1}$ & $ 55.2_{ 6.7}$/ $ 49.0_{ 6.1}$/ $ 51.4_{ 1.4}$\\
\textbf{JNLPBA} & $ 56.9_{ 2.9}$/ $ 67.9_{ 1.7}$/ $ 61.9_{ 2.4}$ & $ 57.4_{ 1.9}$/ $ 73.7_{ 1.8}$/ $ 64.6_{ 1.8}$ & $ 57.2_{ 2.9}$/ $ 75.1_{ 2.4}$/ \textbf{65.0}$ _{ 2.7}$ & $ 44.7_{ 1.0}$/ $ 52.4_{ 3.7}$/ $ 48.3_{ 2.1}$\\
\textbf{BC2GM} & $ 55.4_{ 0.4}$/ $ 57.9_{ 7.2}$/ \textbf{56.5}$ _{ 3.2}$ & $ 53.6_{ 0.8}$/ $ 59.2_{ 2.0}$/ $ 56.2_{ 1.0}$ & $ 49.7_{ 2.1}$/ $ 56.3_{ 5.3}$/ $ 52.7_{ 2.2}$ & $ 43.0_{ 8.2}$/ $ 40.8_{ 2.3}$/ $ 41.4_{ 2.7}$\\ \midrule
\textbf{NER Average} & $ 66.9_{ 1.0}$/ $ 69.9_{ 0.9}$/ \textbf{68.2}$ _{ 0.8}$ & $ 63.7_{ 1.8}$/ $ 71.3_{ 0.4}$/ $ 67.1_{ 0.9}$ & $ 64.0_{ 1.6}$/ $ 71.2_{ 0.5}$/ $ 67.2_{ 0.9}$ & $ 55.1_{ 3.6}$/ $ 49.7_{ 0.6}$/ $ 51.5_{ 1.3}$\\ \midrule \midrule
\textbf{DDI} & $ 19.9_{ 2.0}$/ $ 79.1_{ 3.0}$/ $ 31.8_{ 2.7}$ & $ 17.3_{ 1.4}$/ $ 75.4_{ 1.2}$/ $ 28.2_{ 1.9}$ & $ 25.5_{ 2.2}$/ $ 77.9_{ 3.5}$/ \textbf{38.4}$ _{ 2.6}$ & $ 9.6_{ 1.1}$/ $ 48.6_{ 1.9}$/ $ 16.1_{ 1.6}$ \\
\textbf{ChemProt} & $ 17.9_{ 2.2}$/ $ 62.0_{ 3.9}$/ $ 27.7_{ 2.9}$ & $ 19.0_{ 6.8}$/ $ 60.6_{ 8.2}$/ $ 28.7_{ 8.7}$ & $ 22.0_{ 0.3}$/ $ 69.7_{ 1.2}$/ \textbf{33.4}$ _{ 0.4}$  & $ 15.9_{ 0.8}$/ $ 68.9_{ 1.9}$/ $ 25.9_{ 1.3}$\\
\textbf{GAD} & $ 63.7_{ 6.6}$/ $ 57.2_{ 7.9}$/ $ 60.2_{ 7.4}$ & $ 63.2_{ 5.8}$/ $ 72.7_{ 5.7}$/ $ 67.6_{ 5.8}$ & $ 64.1_{ 4.0}$/ $ 78.5_{ 11.5}$/ \textbf{70.3}$ _{ 5.6}$ & $ 51.4_{ 0.9}$/ $ 92.3_{ 4.2}$/ $ 66.0_{ 1.8}$\\ \midrule
\textbf{RE Average} & $ 33.8_{ 2.0}$/ $ 66.1_{ 2.8}$/ $ 39.9_{ 2.2}$ & $ 33.2_{ 0.6}$/ $ 69.6_{ 2.3}$/ $ 41.5_{ 1.4}$ & $ 37.2_{ 1.8}$/ $ 75.4_{ 4.5}$/ \textbf{47.4}$_{ 1.9}$ & $ 25.6_{ 0.1}$/ $ 70.0_{ 1.4}$/ $ 36.0_{ 0.4}$\\ \bottomrule
\end{tabular}}
\caption{Comparison of the true few-shot performance of fine-tuned BERT-sized PLMs with GPT-3 in-context learning on biomedical IE datasets from the BLURB benchmark \cite{Gu2022DomainSpecificLM}. We run all experiments on at most \num{1000} test examples from each dataset and use \num{3} different \num{100}-example training sets to account for data variance (standard deviation found in subscripts).} 
\label{tab:master}
\vspace{-15pt}
\end{table*}

\subsubsection{Relation Extraction}
\noindent \textbf{DDI.}
The DDI dataset \cite{ddi} consists of sentences from MEDLINE and DrugBank labeled with drug-drug interactions categorized into \num{4} true and one vacuous relation.\\
\noindent \textbf{ChemProt.} ChemProt \cite{chemprot} is a dataset consisting of \num{1820} PubMed abstracts with annotated chemical-protein interactions categorized into \num{5} true and one vacuous relation. \\
\noindent \textbf{GAD.}
The Genetic Association Database corpus \cite{gad} consists of scientific excerpts and abstracts distantly annotated with gene-disease associations.

\subsection{Compared Methods}
In our main experiments, we compare three pre-trained language models, PubMedBERT-base \cite{Gu2022DomainSpecificLM},\footnote{We use the base version of PubMedBERT since larger versions are not publicly available.} BioBERT-large \cite{Lee2020BioBERTAP} and RoBERTa-large \cite{Liu2019RoBERTaAR}, fine-tuned on \num{100} training examples, with GPT-3 in-context learning where each test example's in-context prompt was retrieved from the same \num{100} training examples.\footnote{We used the original \texttt{davinci} model for all GPT-3 experiments.} Both PubMedBERT and BioBERT were pre-trained on a large corpus of PubMed articles; PubMedBERT was pre-trained from scratch with a biomedical-specific vocabulary while BioBERT was initialized from a BERT checkpoint. We use RoBERTa-large as a strong representative for general-domain PLMs. We refer the interested reader to Appendix~\ref{sec:appendix-base-models} for results on the base versions of BioBERT and RoBERTa.

\noindent \textbf{Implementation Details.} We choose \num{100} training examples for our experiments as a reasonable number of annotated examples with which to start training an IE model for a new task.\footnote{A smaller training size (e.g., \num{10}) would likely work in GPT-3's favor but is less representative of practical applications: a serious practitioner would likely annotate \num{100} (compared to \num{10}) examples if it would bring significant gains.} 
For the RE tasks, we use a balanced set of \num{100} examples evenly distributed over all relation types. All BERT-sized PLMs are fine-tuned using the HuggingFace Transformers library \cite{wolf-etal-2020-transformers-custom}. For our GPT-3 experiments, we use a maximum of \num{10} and \num{5} in-context examples for NER and RE respectively to remain within GPT-3's input length limit. Due to the high cost of GPT-3, we evaluate all methods on at most \num{1000} test examples from each dataset, using the same subset for all methods. For RE, the test examples are sampled in a stratified fashion to reflect the original test set distribution of relation types. Model and prompt design selection are done following the true few-shot framework we described in \S\ref{sec:true-few-shot}. To account for training data variance, we run all experiments using \num{3} different \num{100}-example training sets and report the mean and standard deviation.

\section{Results \& Discussion}

\subsection{Main Results}

Our main experimental results can be found in Table \ref{tab:master}. We first note that fine-tuned BERT-sized PLMs outperform GPT-3 in-context learning across all datasets, often by a large margin (on average \num{15.6}-\num{16.7}\% for NER and \num{3.9}-\num{11.4}\% for RE in F1). For NER, even though GPT-3's precision already drops by an average of \num{10} points, recall drops by twice as much. This indicates that entity \textit{underprediction} is an important factor for GPT-3's poor in-context learning performance. In contrast, GPT-3's precision decreases much more steeply in the RE tasks due in part to the poor performance on the \texttt{none} relation class. In \S\ref{sec:error_analysis}, we explore the reasons behind these issues in greater depth.

Drilling down into the fine-tuning results, we note that BERT-sized PLMs obtain reasonable performance on the NER tasks, considering the extremely small size of the training sets. We obtain strong performance in the mid 80s for the drug extraction task (BC5CDR-chem) due to the high lexical regularity of drug names (e.g., suffixes like ``-ate'', ``-ine'' or ``-ol'' are very frequent). On other biomedical NER datasets such as disease and gene extraction, performance stalls in the high and low 60s, respectively. This performance gap is likely due to the higher lexical diversity present in gene and disease names and is also observed in PLMs fine-tuned on the full training sets, which typically achieve scores in the low or mid 80s compared to low 90s for disease recognition~\cite{Gu2022DomainSpecificLM}. It is also worth noting that the base version of PubMedBERT outperforms the larger versions of the general-domain RoBERTa model and biomedicine-specific BioBERT model, suggesting that pre-training on domain-specific text and vocabulary from scratch is especially beneficial for NER, reinforcing the findings in \citet{Gu2022DomainSpecificLM}.

Given the higher complexity of the task, it is not surprising that performance deteriorates for all evaluated methods on RE tasks (especially for DDI and ChemProt since they contain more relation types and higher class imbalance). In contrast with the NER task and previous work using larger training sets \cite{Gu2022DomainSpecificLM}, RoBERTa-large notably outperforms PubMedBERT-base and BioBERT-large in the RE task. This suggests that, in the low-resource setting, larger-scale general-domain pre-training offsets the advantage of domain-specific pre-training in tasks which require more advanced syntactic and semantic understanding such as RE.

\subsection{Ablation Studies for GPT-3}

In Tables \ref{tab:ner-ablation} and \ref{tab:re-ablation}, we present ablation studies demonstrating the effectiveness of the techniques used to improve GPT-3 performance. These studies are done on a subset of \num{250} validation examples from one representative dataset for each task. We follow the LOOCV process discussed in \S\ref{sec:prompt_design} and use the same experimental setup as the main experiments with the exception of using only one \num{100}-example training set instead of three.

We ablate the kNN module for both tasks, replacing it with a module which randomly assigns examples from the training set to each test example's in-context prompt. As we can see in both Table \ref{tab:ner-ablation} and \ref{tab:re-ablation}, removing the kNN module reduces GPT-3 in-context learning performance. Performance drops more steeply for RE than NER, indicating that NER is more resilient to different in-context examples. This is to be expected given that there are only a limited number of completions to choose from in the RE task and thus having similar examples (with likely the same class label as the test example) would favorably bias GPT-3 towards predicting that class label. For NER, conversely, the diversity of entities is large and so it is rare that a training sentence would have similar completions to a given test example in the low-resource setting.

\begin{table}[!h]
\small
\centering
\begin{tabular}{lccc}
\toprule
\textbf{} & \multicolumn{1}{l}{\textbf{F1}} & \multicolumn{1}{l}{\textbf{Precision}} & \multicolumn{1}{l}{\textbf{Recall}} \\ \midrule
\textbf{Best Model}                                                               & \num{46.3}                           & \num{42.5}                                  & \num{50.9}                               \\
\textbf{$-$ kNN Module}                                                           & \num{45.3}                           & \num{42.7}                                  & \num{48.2}                               \\
\textbf{$-$ Logit Biases}                                                          & \num{42.6}                           & \num{66.7}                                  & \num{31.3}                               \\
\textbf{\begin{tabular}[c]{@{}l@{}}$-$ Both\end{tabular}} & \num{38.7}                           & \num{60.2}                                  & \num{28.5}                               \\ \bottomrule
\end{tabular}%
\caption{NER ablation study on BC5CDR-disease.}
\label{tab:ner-ablation}
\vspace{-15pt}
\end{table}

\begin{table}[!h]
\centering
\small
\begin{tabular}{@{}lccc@{}}
\toprule
\textbf{}                                                                         & \multicolumn{1}{l}{\textbf{F1}} & \multicolumn{1}{l}{\textbf{Precision}} & \multicolumn{1}{l}{\textbf{Recall}} \\ \midrule
\textbf{Best Model}                                                               & \num{26.1}                           & \num{16.1}                                  & \num{68.0}                               \\
\textbf{$-$ kNN Module}                                                           & \num{18.6}                           & \num{11.5}                                  & \num{48.0}                               \\
\textbf{$-$ Calibration}                                                          & \num{23.6}                           & \num{14.6}                                  & \num{62.0}                               \\
\textbf{\begin{tabular}[c]{@{}l@{}}$-$ Both\end{tabular}} & \num{16.9}                           & \num{10.9}                                  & \num{38.0}                               \\ \bottomrule
\end{tabular}
\caption{RE ablation study on DDI.}
\label{tab:re-ablation}
\vspace{-8pt}
\end{table}

In our NER-specific ablation study, we find that removing the logit bias option leads to a large drop in performance even though precision improves. This boost in precision is due to our post-processing which removes predicted entities that are not in the original sentence and eliminates false positives. However, since invalid entities are generated instead of the valid spans which could be correct, recall drops. When ablating the kNN module and removing the logit bias option, we see an even greater drop, indicating that they are complementary. As for our RE-specific ablation study, removing the calibration module results in a drop in both precision and recall, with or without the kNN module, verifying its effectiveness.

\subsection{Data Efficiency}

\begin{figure}[!t]
     \centering
     \includegraphics[width=\columnwidth]{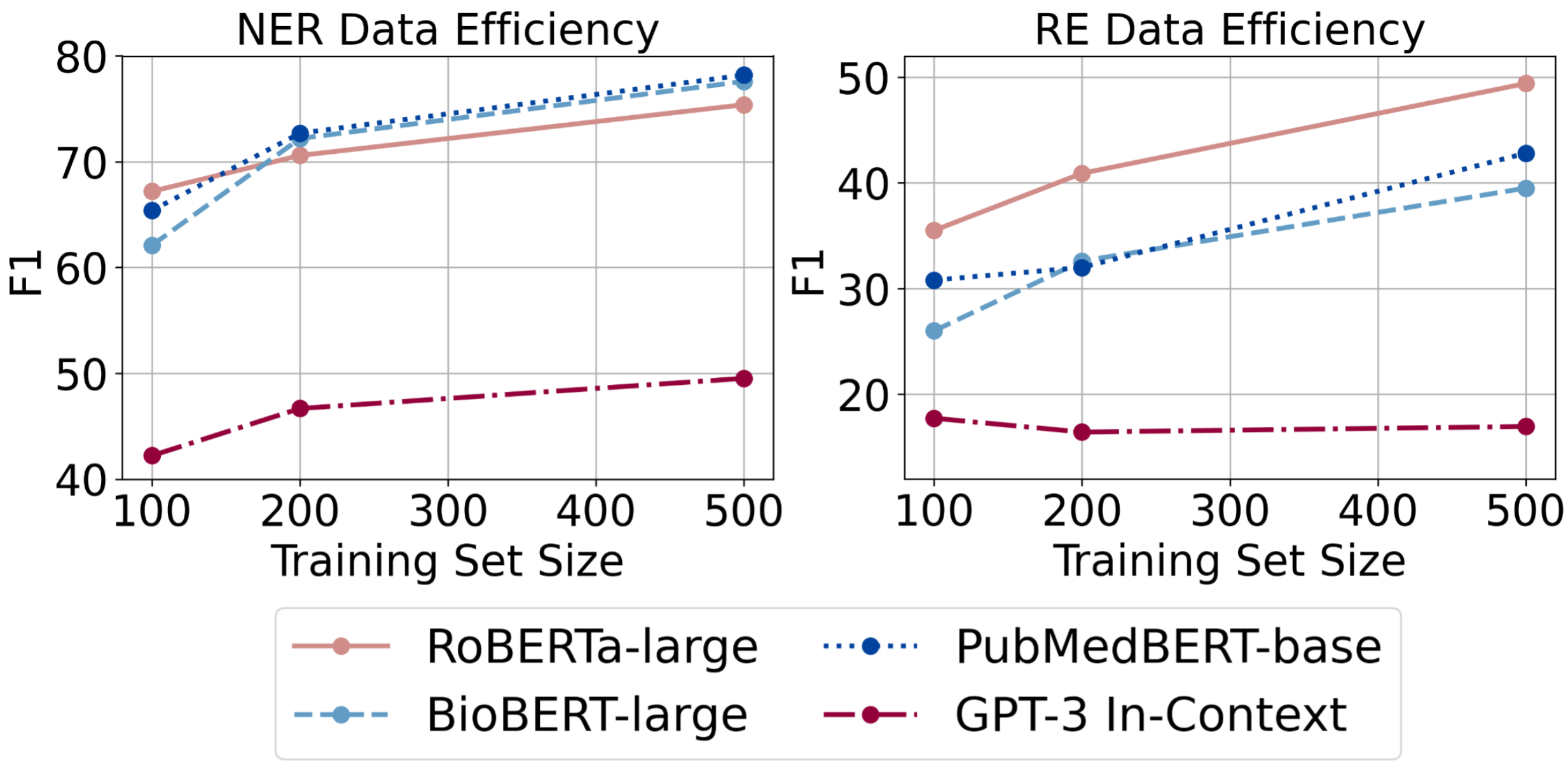}
     \caption{Data efficiency curves for BC5CDR disease NER dataset (left) and DDI RE dataset (right).}
     \label{fig:data_efficiency}
     \vspace{-15pt}
\end{figure}

In practice, choosing an optimal machine learning model requires considering not only a model's overall performance but also, crucially, its \textit{data efficiency}, i.e., how performance improves w.r.t the amount of labeled data added. Previous work shows that GPT-3 in-context learning performance improves as dataset size increases when using kNN retrieval \cite{liu-etal-2022-makes}. Thus, we explore whether adding more training examples to sample from leads to performance improvements via more relevant in-context examples. In this experiment, we expand the training dataset to \num{200} and \num{500} training examples for one representative dataset from each task: BC5CDR-disease and DDI. For the BERT-sized PLMs, we carry out the same cross-validation procedure for model selection as in the main experiments. For GPT-3, we utilize the same optimal prompt design obtained from the main experiments to keep costs manageable. As shown in Figure~\ref{fig:data_efficiency}, for NER, we find that performance for in-context learning improves at a similar rate as the small PLMs, keeping the large gap between them constant. On the other hand, for RE, GPT-3's performance quickly falls behind. This behavior can be partially explained by the fact that \texttt{none} relation examples are more challenging to retrieve by leveraging simple lexical features than their positive class counterparts.\footnote{See \S\ref{sec:re_error_analysis} for more discussion.} Overall, GPT-3 in-context learning does not seem to yield a high return for more data annotation to compensate for its lower few-shot performance, so fine-tuning BERT-sized PLMs is likely still a better choice in the medium to high-data regime.

\subsection{Detailed Error Analysis}\label{sec:error_analysis}

In this section, our in-depth analysis reveals the difficulty of in-context learning in handling the \texttt{null} class,\footnote{It is named after the null hypothesis for its similar nature.} such as sentences that contain no entities (for NER) and entity pairs that hold none of the target relations (for RE). Such issues do not seem to be specific to biomedical applications but are likely detrimental for IE tasks in general.

\subsubsection{NER Error Analysis}

When applying an NER (or similar span extraction tasks such as slot filling) model in practice on an input sentence, it may, more often than not, contain no relevant entity at all (what we call \texttt{null} class examples). For example, up to \num{50}\% sentences in the BC5CDR-disease dataset contain no disease. However, existing work on GPT-3 in-context learning has ignored this issue. For instance, \citet{Zhao2021CalibrateBU} chose to remove all examples that contain no relevant slots from their slot filling experiment. Unfortunately, as we will show, such \texttt{null} class examples turn out to be a major culprit of in-context learning's poor performance.

\begin{table}[!h]
\small
\centering
\resizebox{.9\columnwidth}{!}{%
\begin{tabular}{@{}lccc@{}}
\toprule
\multicolumn{4}{c}{\textbf{Original BC5CDR-disease}}                                                                     \\ \midrule
\textbf{}                                                           & \textbf{F1} & \textbf{Precision} & \textbf{Recall} \\ \midrule
\textbf{GPT-3 In-Context} & \num{43.6}       & \num{57.9}             & \num{35.0}           \\
\textbf{RoBERTa-large}                                              & \num{67.7}       & \num{66.9}              & \num{68.7}        \\ \midrule
\multicolumn{4}{c}{\textbf{Modified BC5CDR-disease}}                                                                    \\ \midrule
\textbf{}                                                           & \textbf{F1} & \textbf{Precision} & \textbf{Recall} \\ \midrule
\textbf{GPT-3 In-Context} & \num{59.8} & \num{60.3} & \num{59.3}	          \\
\textbf{RoBERTa-large}                                              & \num{70.4}       & \num{68.0}            & \num{72.9}          \\ \bottomrule
\end{tabular}%
}
\caption{Evaluation on modified BC5CDR-disease where sentences with no disease entity are removed.}
\label{tab:non_empty_ner}
\vspace{-10pt}
\end{table}

To explore the effect of such \texttt{null} examples, we compare GPT-3 in-context learning with fine-tuned RoBERTa-large on a modified BC5CDR-disease dataset in which all sentences containing no disease entities are removed. As shown in Table \ref{tab:non_empty_ner}, recall for GPT-3 improves by around \num{24}\%, compared to only \num{4}\% for RoBERTa-large, indicating that including \texttt{null} examples in a prompt biases GPT-3 much more strongly to predict few entities than adding them to the fine-tuning data.


\begin{table}[!h]
\centering
\resizebox{\columnwidth}{!}{%
\begin{tabular}{@{}lcccc@{}}
\toprule
\begin{tabular}[c]{@{}c@{}}\textbf{Number of}\\\textbf{Entities}\end{tabular} & 
\begin{tabular}[c]{@{}c@{}}\textbf{$\mathbb{P}($\texttt{null}}$)$\\\textbf{2-Shot}\end{tabular} & \begin{tabular}[c]{@{}c@{}}\textbf{$\mathbb{P}($\texttt{null}}$)$\\\textbf{3-Shot}\end{tabular} & \begin{tabular}[c]{@{}c@{}}\textbf{Absolute}\\ \textbf{$\Delta$}\end{tabular} & \begin{tabular}[c]{@{}c@{}}\textbf{\%}\\ \textbf{Increase}\end{tabular}\\ \midrule
\textbf{Zero (\texttt{null})}&\num{19.4} & \num{49.1} &\num{29.7} & \num{153}\%\\
\textbf{One or More} & \num{15.8}       & \num{40.9} & \num{25.1}&  \num{159}\%\\ \bottomrule
\end{tabular}%
}
\caption{We compare the \texttt{null} token probability assigned by GPT-3 to examples with zero and non-zero entities in the BC5CDR-disease training dataset. We run GPT-3 on 2-shot and 3-shot prompts (the 3-shot prompts contain one extra \texttt{null} example to examine its effect). We present the average over 3 randomly chosen prompts.}
\label{tab:ner_error_analysis}
\vspace{-5pt}
\end{table}


We hypothesize that this bias is due, at least in part, to the fact that GPT-3 in-context learning must infer that relevant entities should only be predicted if they are present in the given sentence, in contrast with smaller PLMs using the token-classification formulation. In order to examine this hypothesis more closely, we simplify our experimental setting to isolate the effect that an additional \texttt{null} example has on GPT-3's predictions. We run GPT-3 on the BC5CDR-disease training dataset without the k-NN retrieval module, instead using the same randomly chosen two-shot prompt (containing an example with no entities and one with at least one) across all examples. We then add one more random example without entities to every prompt and compare the probability of a \texttt{null} prediction in each setting.\footnote{We measure \texttt{null} prediction probability instead of performance since entity-level F1 would not capture any information about examples with no entities.} As shown in Table \ref{tab:ner_error_analysis}, we find that, while adding the second \texttt{null} example increases the \texttt{null} probability slightly more for zero entity examples than ones with entities in absolute terms, accounting for the lower initial \texttt{null} probability that is assigned to examples with one entity or more reverses this effect. The absence of a significantly larger increase on the \texttt{null} probability for examples with zero entities over others suggests that GPT-3 struggles to infer the appropriate prediction constraint for this task and rather increases the \texttt{null} probability somewhat uniformly across examples.

\begin{table*}[!t]
\small
\renewcommand{\arraystretch}{1.2}
\resizebox{\textwidth}{!}{%
\begin{tabular}{@{}clcc@{}}
\toprule
\textbf{Label}             & \textbf{Example} & \textbf{Model} & \textbf{Correct} \\\midrule
\multirow{3}{*}{Effect} &  \begin{tabular}[c]{@{}l@{}}Concurrent use of \hlc[cyan!0!white]{\textbf{phenothiazines}} \hlc[cyan!5!white]{may} antagonize \hlc[cyan!10!white]{the} \hlc[cyan!50!white]{anorectic} \hlc[cyan!42!white]{effect} of \hlc[cyan!0!white]{\textbf{diethylpropion}}. \end{tabular} & RoBERTa-large  & \cmark\\ \cmidrule(l){2-4} &  \begin{tabular}[c]{@{}l@{}}\hlc[cyan!22!white]{Concurrent} \hlc[cyan!16!white]{use} \hlc[cyan!24!white]{of} \hlc[cyan!0!white]{\textbf{phenothiazines}} may \hlc[cyan!6!white]{antagonize} the \hlc[cyan!50!white]{anorectic} \hlc[cyan!33!white]{effect} \hlc[cyan!24!white]{of} \hlc[cyan!0!white]{\textbf{diethylpropion}}.\end{tabular} & GPT-3 & \cmark\\\midrule

\multirow{3}{*}{None} & \begin{tabular}[c]{@{}l@{}}Other \hlc[cyan!21!white]{strong} inhibitors \hlc[cyan!50!white]{of} \hlc[cyan!41!white]{CYP3A4} (\hlc[cyan!46!white]{e.g.}, itraconazole, \hlc[cyan!0!white]{\textbf{clarithromycin}}, \hlc[cyan!1!white]{nefazodone}, troleandomycin,\\ ritonavir, \hlc[cyan!0!white]{\textbf{nelfinavir}}) would be expected to behave similarly. \end{tabular} & RoBERTa-large  & \cmark\\ \cmidrule(l){2-4} &  \begin{tabular}[c]{@{}l@{}}Other strong inhibitors of CYP3A4 (e.g., itraconazole, \hlc[cyan!0!white]{\textbf{clarithromycin}}, nefazodone, troleandomycin, \\ritonavir, \hlc[cyan!0!white]{\textbf{nelfinavir}}) would be \hlc[cyan!23!white]{expected} \hlc[cyan!39!white]{to} \hlc[cyan!50!white]{behave} \hlc[cyan!35!white]{similarly}.\end{tabular} & GPT-3 & \begin{tabular}[c]{@{}c@{}}\xmark\\(Mechanism)\end{tabular}\\\bottomrule
\end{tabular}%
}
\caption{We compare LIME-based saliency scores for two DDI examples predicted by GPT-3 in-context learning and RoBERTa-large. Masking out words highlighted in \textcolor{cyan}{blue} changes the model's current prediction (the color's intensity indicates the effect of removing each word on the final prediction). The drugs shown in \textbf{bold} are the head and tail entities for the relation being queried. The second example shows that GPT-3 in-context learning is more prone to spurious surface-level signals and thus suffers in correctly predicting the \texttt{none} class.}
\label{tbl:re_error_analysis}
\vspace{-12pt}
\end{table*}

\subsubsection{RE Error Analysis}\label{sec:re_error_analysis}

We similarly examine the effect of the \texttt{null} class for RE, which is denoted as the \texttt{none}
relation in the DDI dataset analyzed. As seen in Table \ref{tab:master}, GPT-3 in-context learning achieves high recall but low precision on RE datasets that have multiple relation types such as DDI and ChemProt. Based on the confusion matrices derived from LOOCV (Appendix~\ref{sec:re_confusion_analysis}), the \texttt{none} relation in DDI is rarely predicted by GPT-3. This bias against the \texttt{none} class greatly degrades the model's precision given that the DDI dataset is, rightfully so, heavily skewed towards this class.

In order to further understand this bias, we use LIME \cite{lime}\footnote{Our LIME process is described in Appendix~\ref{sec:lime}.} to analyze the predictions for both GPT-3 and RoBERTa on an \texttt{effect} example and a \texttt{none} example.\footnote{Other similar examples are discussed in Appendix~\ref{sec:qualitative}.} The first example in Table \ref{tbl:re_error_analysis} was labeled correctly by both models by relying on "anorectic effect" as a relevant signal. For \texttt{none} examples, however, correct predictions often require the use of more implicit structural understanding rather than reliance on surface level signals, as can be seen in the second example in Table \ref{tbl:re_error_analysis}. In this \texttt{none} example, we note that RoBERTa-large's prediction is strongly affected by the phrase ``\textit{of CYP3A4 (e.g.,}'' which helps express that the drugs within the parenthesis are examples of the same drug class and therefore do not interact with each other. This suggests that RoBERTa correctly leverages the linguistic structure of the sentence. On the other hand, GPT-3's incorrect \texttt{mechanism} prediction appears to be supported by the phrase ``\textit{expected to behave similarly}'', which is not relevant to the relation between the drugs being queried. This suggest that GPT-3 in-context learning is more prone to spurious surface-level signals and thus suffers in predicting the \texttt{none} class.

\subsubsection{General Limitation or Domain Shift?}

Our analysis suggests that GPT-3's in-context learning faces a broader issue concerning the higher complexity of \texttt{null} examples compared to positive examples. However, given that there is little work thoroughly studying GPT-3 for general domain IE, we leave it for future work to determine to what extent our findings stem from this \texttt{null} class limitation, the biomedical domain shift, or some other unforeseen reasons.

\section{Related Work}

\noindent \textbf{In-Context Learning.}
GPT-3 in-context learning \cite{Brown2020LanguageMA} has been found to be competitive against supervised baselines in a broad range of tasks including text classification, natural language inference, machine translation, question answering, table-to-text generation and semantic parsing \cite{Brown2020LanguageMA, Zhao2021CalibrateBU, liu-etal-2022-makes, shin-etal-2021-constrained}. Many techniques have been introduced to bolster its performance by removing biases through calibration  \cite{Zhao2021CalibrateBU, malkin2021boosting} as well as by optimizing prompt retrieval \cite{liu-etal-2022-makes, rubin-etal-2022-learning, shin-etal-2021-constrained}, prompt ordering \cite{lu-etal-2022-fantastically} and prompt design \cite{Perez2021TrueFL}.

Previous work exploring GPT-3's in-context learning performance for information extraction tasks is limited. \citet{Zhao2021CalibrateBU} evaluate smaller GPT-3 models on a modified slot filling task in which all examples have at least one entity of interest. Additionally, \citet{Epure2021ARS} evaluate the in-context learning performance of GPT-2 on open-domain NER datasets, modified to keep a specific ratio of empty to non-empty examples. Our prompt design for biomedical NER draws heavily from both of these works. As far as we know, our work is among the first to comprehensively evaluate GPT-3's in-context learning performance on IE tasks.\\

\noindent \textbf{Prompt Design.}
Apart from work on in-context learning, several other research directions study how to reformulate NLP tasks as language generation tasks. \citet{schick-schutze-2021-just} reformulate text classification and natural language inference tasks using a diverse set of manually constructed cloze-style templates as prompts to improve few-shot learning in smaller pretrained language models. \citet{Gao2021MakingPL} explore a similar setting but leverage an external language model to generate such templates. Both of these demonstrate the importance of using a variety of prompt designs.

In a related direction, \citet{huguet-cabot-navigli-2021-rebel-relation} achieve state-of-the-art performance on relation extraction benchmarks by reformulating it as an end-to-end sequence-to-sequence task. In the biomedical domain, several works \cite{Raval2021ExploringAU, Phan2021SciFiveAT, parmar-etal-2022-boxbart} follow the multi-task sequence-to-sequence paradigm introduced by \citet{Raffel2020ExploringTL} and outperform previous methods on many tasks such as side effect extraction, NER, RE, natural language inference and question answering. Our prompt design is heavily inspired by many of these efforts to reformulate IE tasks as sequence-to-sequence tasks.\\

\noindent \textbf{True Few-Shot Learning.}
\citet{Perez2021TrueFL} argue that previous work overestimates the few-shot learning abilities of PLMs by using large validation sets for model and prompt selection. This setting has been adopted by many works in this direction in an effort to more accurately estimate few-shot performance \cite{logan-iv-etal-2022-cutting, schick-schutze-2022-true, lu-etal-2022-fantastically}.\\

\noindent \textbf{Biomedical In-Context Learning.}
Previous work evaluating GPT-3's in-context learning abilities on biomedical NLP tasks suggests that using the GPT-3 API directly yields poor performance in the biomedical domain \cite{Moradi2021GPT3MA}. Their work provides experimental results on \num{5} biomedical NLP datasets on distinct tasks including relation extraction. In our study, we aim to provide a comprehensive and in-depth evaluation on biomedical IE by using an established multi-dataset biomedical NLP benchmark and leverage recent in-context learning techniques to obtain the highest possible performance to our knowledge and ability. However, our results ultimately provide more evidence for the inadequacy of GPT-3 in-context learning for biomedical IE tasks, which cannot be easily overcome with existing techniques. Interestingly, a concurrent work \cite{zero-shot-clinical} finds that GPT-3 perform well on a different set of \textit{clinical} IE tasks, including one on biomedical evidence extraction that is clinical in nature. More work is needed to ascertain the cause of this surprising gap in IE performance between the clinical and biomedical domains for in-context learning.

\section{Conclusions}

In this work, we explored the potential of GPT-3 in-context learning for the high impact task of biomedical information extraction (IE). Given that such a paradigm would provide significant advantages for biomedical IE applications, we spent considerable efforts exploring available techniques that have been proven effective for other in-context learning settings. We showed, however, that current techniques do not enable GPT-3 in-context learning to surpass BERT-sized PLM fine-tuning on a range of benchmark datasets for biomedical NER and RE. Additionally, we discussed some potentially general limitations of in-context learning in biomedical IE to be explored in future work: its difficulty in handling the \texttt{null} class, such as entity-less NER examples and vacuous relation examples for RE. Apart from posing this question for further study, we hope our work provides helpful guidance to biomedical researchers and practitioners towards more promising and cost-effective tools for low-resource IE such as small PLM fine-tuning or perhaps even directly fine-tuning GPT-3.

\section*{Limitations}

While we have uncovered a large performance gap between current GPT-3 in-context learning techniques and standard fine-tuning in the true few-shot setting, there are several important limitations that are worth discussing. Our limited budget restricted our study to a small set of prompt styles and number of examples in the prompt. Although our experiments suggest otherwise, it is possible that having a larger prompt design search space or using more examples per prompt could narrow the gap between small PLM fine-tuning and GPT-3 in-context learning. Additionally, it is still unclear to what degree using larger validation sets, at the cost of compromising the few-shot assumption, for prompt selection could improve GPT-3's in-context learning performance. Perhaps more notably, the kNN retrieval module used in this study relies on whole sentence embeddings, as commonly done in the existing literature. However, intuitively, tasks like relation extraction require a more focused view around the target entity pair. We speculate that developing a better retrieval module that is able to incorporate such task-specific inductive biases will likely be beneficial for in-context learning, but we leave it for future work. 
Finally, it is important to note that while contextual calibration \cite{Zhao2021CalibrateBU} is shown to work well in some text classification tasks, it is unclear whether other more recent methods such as that by \citet{malkin2021boosting} could better address GPT-3's text generation biases or if more task-specific calibration mechanisms are necessary for IE tasks.

\section*{Acknowledgements}
The authors would like to thank colleagues from the OSU NLP group for their thoughtful comments. This research was supported in part by OSU TDAI, NSF OAC 2118240, NSF OAC 2112606, NSF IIS 1815674, NSF CAREER 1942980, and Ohio Supercomputer Center \cite{OhioSupercomputerCenter1987}.

\bibliography{anthology, custom}
\bibliographystyle{acl_natbib}

\clearpage
\appendix

\section{Experimental Setup Details}

\subsection{Named Entity Recognition}

We follow the BIO tag formulation for NER and use standard fine-tuning process for PLMs used in  \citet{Gu2022DomainSpecificLM}. Given a sentence containing $n$ tokens $X=[x_1, ..., x_n]$, an NER system attempts to predict a tag for each token: $Y=[y_1, ...,y_n]$, which can then be translated into a set of $k$ entities. An encoder $H$ is used to obtain a contextualized representation for the sentence $X$: $H(X) = [\vec{v_1}, ..., \vec{v_n}]$. Each embedding $\vec{v_i}$ is then used to predict $y_i$ using a linear layer. The encoder $H$ and the linear layer are then fine-tuned using a standard cross entropy objective. We use NLTK \cite{Bird2010BookRN} to tokenize all NER sentences.

\subsection{Relation Extraction}

For RE we use the simplest formulation in standard fine-tuning, the subject and object entities for each relation are replaced in the original sentence by new special tokens [ENT1] and [ENT2]. An encoder $H$ is then used as in NER to obtain a contextualized representation $H(X) = [\vec{v_1}, ..., \vec{v_n}]$ of the now masked sentence. As is standard for text classification tasks, the [CLS] token embedding is then used to predict each relation type. As with the NER task, a standard cross entropy loss is used to fine-tune the encoder and linear layer.

\section{Computational Cost}

For our experiments, we used \num{4} NVIDIA GeForce RTX 2080 Ti GPUs. The number of parameters for each model we used as well as the total GPU hours and costs associated with using GPT-3 are listed in Table \ref{tab:computational_costs}.

\begin{table}[!h]
\centering
\resizebox{\columnwidth}{!}{%
\begin{tabular}{@{}lccc@{}}
\toprule
\multicolumn{1}{c}{}             & \textbf{\begin{tabular}[c]{@{}c@{}}\# of Parameters\\ (millions)\end{tabular}} & \textbf{\begin{tabular}[c]{@{}c@{}}Total GPU \\ Hours\end{tabular}} & \textbf{\begin{tabular}[c]{@{}c@{}}Total \\ Cost\end{tabular}} \\ \midrule
\textbf{RoBERTa-large}           & \num{  354                                                                             } & \num{  338                                                              } & -                                                              \\
\textbf{PubMedBERT-base}         & \num{  100                                                                             } & \num{  138                                                              } & -                                                              \\
\textbf{BioBERT-large}           & \num{  345                                                                             } & \num{  344                                                              } & -                                                              \\
\textbf{GPT-3 (davinci)} & \num{  175000                                                                           } & -                                                                   & $\sim$\$\num{1500}                                                  \\ \bottomrule
\end{tabular}%
}
\caption{Total GPU Hours and GPT-3 costs associated with our experiments.}
\label{tab:computational_costs}
\end{table}

\newpage

\section{Fine-Tuning Hyperparameters}\label{sec:appendix-hps}

We run 5-fold cross validation for each \num{100} sample training subset to choose between all hyperparameters listed in Table \ref{tab:hyperparameter_search_grid}.

\begin{table}[!h]
\centering
\small
\resizebox{\columnwidth}{!}{%
\begin{tabular}{@{}lccccc@{}}
\toprule
 & \multicolumn{1}{c}{\textbf{\begin{tabular}[c]{@{}c@{}}Learning \\Rate\end{tabular}}}                       & \multicolumn{1}{c}{\textbf{\begin{tabular}[c]{@{}c@{}}Batch\\ Size\end{tabular}}}        & \multicolumn{1}{c}{\textbf{\begin{tabular}[c]{@{}c@{}}Warmup \\Ratio\end{tabular}}}         & \multicolumn{1}{c}{\textbf{\begin{tabular}[c]{@{}c@{}}Weight \\Decay\end{tabular}}}               & \multicolumn{1}{c}{\textbf{\begin{tabular}[c]{@{}c@{}}Early Stopping\\ Checkpoint\end{tabular}}} \\ \midrule
\textbf{\begin{tabular}[c]{@{}l@{}}Search\\ Space\end{tabular}}  & \begin{tabular}[c]{@{}c@{}}\num{1}e-\num{5}\\ \num{2}e-\num{5}\\ \num{3}e-\num{5}\\ \num{5}e-\num{5}\end{tabular} & \begin{tabular}[c]{@{}c@{}}\num{16}\\ \num{32}\end{tabular} & \begin{tabular}[c]{@{}c@{}}\num{0.0}\\ \num{0.06}\end{tabular} & \begin{tabular}[c]{@{}c@{}}\num{0.0}\\ \num{0.01}\\ \num{0.1}\end{tabular} & \begin{tabular}[c]{@{}c@{}}\num{5}\\ \num{10}\\ \num{15}\\ \num{20}\\ \num{25}\end{tabular}                                    \\ \bottomrule
\end{tabular}%
}
\caption{Hyperparameter search grid used with k-fold cross-validation to obtain the optimal hyperparameters for all PLM fine-tuning experiments.}
\label{tab:hyperparameter_search_grid}
\end{table}

\section{Prompt Designs}\label{sec:appendix-prompts}

We run leave-one-out cross validation for each 100 sample training subset to choose between all choices listed in Table \ref{tab:prompt_designs_params}. Prompt design selections  were completely independent for each training subset to maintain the true few-shot learning setting.

\begin{table*}[!p]
\centering
\resizebox{\textwidth}{!}{%
\begin{tabular}{llllcc}                                                              \hline \\[-8pt]
\multicolumn{6}{c}{\textbf{NER}}                                                                                                                                                                                                                                                                                                                                                                                                                                                                                                                                                                                                                                                                                                                                                           \\[5pt] \hline \\[-8pt]
\multicolumn{1}{c}{}                      & \multicolumn{1}{c}{\textbf{\begin{tabular}[c]{@{}c@{}}Overall\\ Instructions\end{tabular}}}                                                                                        & \multicolumn{1}{c}{\textbf{\begin{tabular}[c]{@{}c@{}}Sentence\\ Introduction\end{tabular}}} & \multicolumn{1}{c}{\textbf{\begin{tabular}[c]{@{}c@{}}Retrieval\\ Message\end{tabular}}}                                                                                                                                                                                                         & \textbf{\begin{tabular}[c]{@{}c@{}}\# of \\ In-Context \\ Examples\end{tabular}} & \multicolumn{1}{c}{\textbf{Label Verbalizer}}                                                                         \\[20pt] \hline \\[-8pt]                   
\multirow{3}{*}{\textbf{BC5CDR-disease}}  & ""                                                                                                                                                                                 & Sentence:                                                                                    & \multicolumn{1}{c}{\multirow{3}{*}{Diseases:}}                                                                                                                                                                                                                                                                       & 5                                                                             & \multirow{15}{*}{N/A}                                                                                                                                                                                                                                          \\[5pt] \cline{2-3} \cline{5-5} \\[-8pt]
                                          & \begin{tabular}[c]{@{}l@{}}List the diseases mentioned\\ in the following sentences.\end{tabular}                                                                                  & \begin{tabular}[c]{@{}l@{}}Scientific \\ Article Excerpt:\end{tabular}                       &                                                                                                                                                                                                                                                                                                  & 10                                                                            &                                                                                                                                                                                                                                                                \\[10pt] \cline{1-5} \\[-8pt]
\multirow{3}{*}{\textbf{BC5CDR-chemical}} & ""                                                                                                                                                                                 & Sentence:                                                                                    & \multicolumn{1}{c}{\multirow{3}{*}{Drugs:}}                                                                                                                                                                                                                                                                & 5                                                                             &                                                                                                                                                                                                                                                                \\[5pt] \cline{2-3} \cline{5-5} \\[-8pt]
                                          & \begin{tabular}[c]{@{}l@{}}List the drugs mentioned\\ in the following sentences.\end{tabular}                                                                                     & \begin{tabular}[c]{@{}l@{}}Scientific \\ Article Excerpt:\end{tabular}                       &                                                                                                                                                                                                                                                                                                  & 10                                                                            &                                                                                                                                                                                                                                                                \\[10pt] \cline{1-5}\\[-8pt]
\multirow{3}{*}{\textbf{NCBI-disease}}    & ""                                                                                                                                                                                 & Sentence:                                                                                    & \multicolumn{1}{c}{\multirow{3}{*}{Diseases:}}                                                                                                                                                                                                                                                                       & 5                                                                             &                                                                                                                                                                                                                                                                \\[5pt] \cline{2-3} \cline{5-5}\\[-8pt]
                                          & \begin{tabular}[c]{@{}l@{}}List the diseases mentioned\\ in the following sentences.\end{tabular}                                                                                  & \begin{tabular}[c]{@{}l@{}}Scientific \\ Article Excerpt:\end{tabular}                       &                                                                                                                                                                                                                                                                                                  & 10                                                                            &                                                                                                                                                                                                                                                                \\[10pt] \cline{1-5}\\[-8pt]
\multirow{3}{*}{\textbf{JNLPBA}}          & ""                                                                                                                                                                                 & Sentence:                                                                                    & \multicolumn{1}{c}{\multirow{3}{*}{Genes:}}                                                                                                                                                                                                                                                                          & 5                                                                             &                                                                                                                                                                                                                                                                \\[5pt] \cline{2-3} \cline{5-5}\\[-8pt]
                                          & \begin{tabular}[c]{@{}l@{}}List the genes mentioned \\ in the following sentences.\end{tabular}                                                                                    & \begin{tabular}[c]{@{}l@{}}Scientific \\ Article Excerpt:\end{tabular}                       &                                                                                                                                                                                                                                                                                                  & 10                                                                            &                                                                                                                                                                                                                                                                \\[10pt] \cline{1-5}\\[-8pt]
\multirow{3}{*}{\textbf{BC2GM}}           & ""                                                                                                                                                                                 & Sentence:                                                                                    & \multicolumn{1}{c}{\multirow{3}{*}{Genes:}}                                                                                                                                                                                                                                                                          & 5                                                                             &                                                                                                                                                                                                                                                                \\[5pt] \cline{2-3} \cline{5-5}\\[-8pt]
                                          & \begin{tabular}[c]{@{}l@{}}List the genes mentioned \\ in the following sentences.\end{tabular}                                                                                    & \begin{tabular}[c]{@{}l@{}}Scientific \\ Article Excerpt:\end{tabular}                       &                                                                                                                                                                                                                                                                                                  & 10                                                                            &                                                                                                                                                                                                                                                                \\[20pt] \hline \\[-8pt]
\multicolumn{6}{c}{\textbf{RE}}                                                                                                                                                                                                                                                                                                                                                                                                                                                                                                                                                                                                                                                                                                                                                           \\[5pt] \hline \\[-8pt]
\multicolumn{1}{c}{}                      & \multicolumn{1}{c}{\textbf{\begin{tabular}[c]{@{}c@{}}Overall\\ Instructions\end{tabular}}}                                                                                        & \multicolumn{1}{c}{\textbf{\begin{tabular}[c]{@{}c@{}}Sentence\\ Introduction\end{tabular}}} & \multicolumn{1}{c}{\textbf{\begin{tabular}[c]{@{}c@{}}Retrieval\\ Message\end{tabular}}}                                                                                                                                                                                                         & \textbf{\begin{tabular}[c]{@{}c@{}}\# of \\ In-Context \\ Examples\end{tabular}} & \multicolumn{1}{c}{\textbf{Label Verbalizer}}                                                                         \\[20pt] \hline \\[-8pt]  
\multirow{8}{*}{\textbf{DDI}}             & ""                                                                                                                                                                                 & Sentence:                                                                                    & \begin{tabular}[c]{@{}l@{}}Drug 1: \textless{}DRUG1\textgreater\\ Drug 2: \textless{}DRUG2\textgreater\\ Interaction:\end{tabular}                                                                                                                                                               & \multirow{8}{*}{5}                                                            & \multicolumn{1}{l}{\multirow{8}{*}{\begin{tabular}[c]{@{}l@{}}DDI-effect $\textgreater$ effect\\ DDI-false $\textgreater$ none\\ DDI-advise $\textgreater$ advice\\ DDI-mechanism $\textgreater$ mechanism\\ DDI-int $\textgreater$ other\end{tabular}}}                 \\[15pt] \cline{2-4}\\[-8pt]
                                          & \begin{tabular}[c]{@{}l@{}}Classify the interaction \\ between drugs based on\\  the provided scientific \\ article excerpts.\end{tabular}                                         & \begin{tabular}[c]{@{}l@{}}Scientific \\ Article Excerpt:\end{tabular}                       & \begin{tabular}[c]{@{}l@{}}How do \textless{}DRUG1\textgreater \\ and \textless{}DRUG2\textgreater interact \\ according to the previous \\ sentence? Which word \\ best describes their \\ interaction: advice, effect, \\ mechanism, other or none? \\ Interaction:\end{tabular}               &                                                                               & \multicolumn{1}{l}{}                                                                                                                                                                                                                                           \\[52pt] \hline\\[-8pt]
\multirow{8}{*}{\textbf{ChemProt}}        & ""                                                                                                                                                                                 & Sentence:                                                                                    & \begin{tabular}[c]{@{}l@{}}Drug: \textless{}DRUG\textgreater\\ Gene: \textless{}GENE\textgreater\\ Effect:\end{tabular}                                                                                                                                                                          & \multirow{8}{*}{5}                                                            & \multicolumn{1}{l}{\multirow{8}{*}{\begin{tabular}[c]{@{}l@{}}false $\textgreater$ none\\ CPR:3 $\textgreater$ activator\\ CPR:4 $\textgreater$ inhibitor\\ CPR:5 $\textgreater$ agonist\\ CPR:6 $\textgreater$ antagonist\\ CPR:9 $\textgreater$ substrate\end{tabular}}} \\[20pt] \cline{2-4}\\[-8pt]
                                          & \begin{tabular}[c]{@{}l@{}}Classify the effect drugs \\ have on the genes \\ mentioned in the following\\  scientific article excerpts.\end{tabular}                               & \begin{tabular}[c]{@{}l@{}}Scientific \\ Article Excerpt:\end{tabular}                       & \begin{tabular}[c]{@{}l@{}}What effect does the drug \\ \textless{}DRUG\textgreater have on gene\\  \textless{}GENE\textgreater according to the \\ previous sentence? Choose \\ from the following: none, \\ activator, inhibitor, agonist, \\ antagonist or substrate. \\ Effect:\end{tabular} &                                                                               & \multicolumn{1}{l}{}                                                                                                                                                                                                                                           \\[50pt] \hline\\[-8pt]
\multirow{8}{*}{\textbf{GAD}}             & ""                                                                                                                                                                                 & Sentence:                                                                                    & \begin{tabular}[c]{@{}l@{}}Gene: \textless{}GENE\textgreater\\ Disease: \textless{}DISEASE$\textgreater$\\ Interaction:\end{tabular}                                                                                                                                                               & \multirow{8}{*}{5}                                                            & \multicolumn{1}{l}{\multirow{8}{*}{\begin{tabular}[c]{@{}l@{}}0 $\textgreater$ no\\ 1 $\textgreater$ yes\end{tabular}}}                                                                                                                                            \\[20pt] \cline{2-4} \\[-8pt]
                                          & \begin{tabular}[c]{@{}l@{}}Determine if there is \\ any interaction between \\ the diseases and genes \\ mentioned in the \\ provided scientific \\ article excerpts.\end{tabular} & \begin{tabular}[c]{@{}l@{}}Scientific \\ Article Excerpt:\end{tabular}                       & \begin{tabular}[c]{@{}l@{}}Based on the previous sentence, \\ is there any interaction between \\ gene \textless{}GENE\textgreater and disease \\ \textless{}DISEASE\textgreater{}?\end{tabular}                                                                                                 &                                                                               & \multicolumn{1}{l}{}                                                                                                                                                                                                                                           \\[40pt] \hline\\[-8pt]
\end{tabular}%
}
\caption{For each element in our proposed prompt design (overall task instruction, sentence introduction and retrieval message), we list every option used for each dataset. For our main experiments, we used LOOCV on 100 training examples to select among 8 combinations of our 3 design elements and the number of in-context examples added to the prompt for each task. We also list the label verbalization used for each relation extraction dataset.}
\label{tab:prompt_designs_params}
\end{table*}

\begin{table*}[!t]
\small
\centering
\resizebox{\textwidth}{!}{%
\begin{tabular}{@{}lcccccc@{}}
\toprule
 &
  \textbf{PubMedBERT-base} &
  \textbf{BioBERT-base} &
  \textbf{RoBERTa-base} &
  \textbf{BioBERT-large} &
  \textbf{RoBERTa-large} &
  \textbf{\begin{tabular}[c]{@{}c@{}}GPT-3 In-Context\end{tabular}} \\\cmidrule(l){2-7}
   & Precision / Recall / F1 & Precision / Recall / F1 & Precision / Recall / F1 & Precision / Recall / F1 & Precision / Recall / F1 & Precision / Recall / F1  \\\midrule
\textbf{BC5CDR-disease} & $67.4_{ 3.7}$/ $67.5_{ 1.2}$/ $67.4_{ 2.4}$ & $60.6_{ 5.1}$/ $66.1_{ 5.7}$/ $63.0_{ 2.0}$ & $60.4_{ 2.8}$/ $61.9_{ 4.4}$/ $61.2_{ 3.6}$ & $62.9_{ 5.0}$/ $69.0_{ 3.0}$/ $65.8_{ 4.1}$ & $66.9_{ 1.7}$/ $68.7_{ 4.7}$/ \textbf{67.7}$ _{ 1.8}$  & $ 57.9_{ 2.3}$/ $ 35.0_{ 2.9}$/ $ 43.6_{ 2.2}$\\
\textbf{BC5CDR-chem} & $86.1_{ 1.9}$/ $88.6_{ 4.8}$/ \textbf{87.3}$_{ 1.3}$ & $77.8_{ 3.3}$/ $85.1_{ 4.6}$/ $81.2_{ 0.5}$ & $74.6_{ 3.8}$/ $84.1_{ 2.1}$/ $79.0_{ 1.2}$ & $84.8_{ 2.6}$/ $87.3_{ 3.3}$/ $86.0_{ 1.1}$ & $82.1_{ 1.8}$/ $87.3_{ 1.0}$/ $84.6_{ 1.3}$ & $ 74.7_{ 2.5}$/ $ 71.4_{ 2.2}$/ $ 73.0_{ 0.3}$\\
\textbf{NCBI-disease} & $68.5_{ 4.7}$/ $67.6_{ 2.4}$/ \textbf{68.0}$_{ 2.9}$ & $58.8_{ 5.4}$/ $65.9_{ 2.7}$/ $62.1_{ 4.0}$ & $60.6_{ 3.2}$/ $61.9_{ 4.6}$/ $61.2_{ 3.5}$ & $59.6_{ 10.6}$/ $67.0_{ 6.1}$/ $63.0_{ 8.7}$ & $64.3_{ 3.7}$/ $68.7_{ 6.7}$/ $66.4_{ 5.1}$ & $ 55.2_{ 6.7}$/ $ 49.0_{ 6.1}$/ $ 51.4_{ 1.4}$\\
\textbf{JNLPBA} & $56.9_{ 2.9}$/ $67.9_{ 1.7}$/ $61.9_{ 2.4}$ & $49.1_{ 0.2}$/ $66.7_{ 1.9}$/ $56.6_{ 0.8}$ & $54.6_{ 2.7}$/ $71.4_{ 2.6}$/ $61.9_{ 2.7}$ & $57.4_{ 1.9}$/ $73.7_{ 1.8}$/ $64.6_{ 1.8}$ & $57.2_{ 2.9}$/ $75.1_{ 2.4}$/ \textbf{65.0}$_{ 2.7}$ & $ 44.7_{ 1.0}$/ $ 52.4_{ 3.7}$/ $ 48.3_{ 2.1}$\\
\textbf{BC2GM} & $55.4_{ 0.4}$/ $57.9_{ 7.2}$/ \textbf{56.5}$_{ 3.2}$ & $46.4_{ 2.5}$/ $57.3_{ 1.0}$/ $51.3_{ 1.9}$ & $46.2_{ 3.0}$/ $53.7_{ 0.4}$/ $49.7_{ 1.6}$ & $53.6_{ 0.8}$/ $59.2_{ 2.0}$/ $56.2_{ 1.0}$ & $49.7_{ 2.1}$/ $56.3_{ 5.3}$/ $52.7_{ 2.2}$ & $ 43.0_{ 8.2}$/ $ 40.8_{ 2.3}$/ $ 41.4_{ 2.7}$\\\midrule
\textbf{NER Average} & $66.9_{ 1.0}$/ $69.9_{ 0.9}$/ \textbf{68.2}$_{ 0.8}$ & $58.6_{ 0.9}$/ $68.2_{ 1.6}$/ $62.8_{ 1.0}$ & $59.3_{ 2.8}$/ $66.6_{ 1.7}$/ $62.6_{ 2.2}$ & $63.7_{ 1.8}$/ $71.3_{ 0.4}$/ $67.1_{ 0.9}$ & $64.0_{ 1.6}$/ $71.2_{ 0.5}$/ $67.2_{ 0.9}$ & $ 55.1_{ 3.6}$/ $ 49.7_{ 0.6}$/ $ 51.5_{ 1.3}$\\\midrule \midrule
\textbf{DDI} & $19.9_{ 2.0}$/ $79.1_{ 3.0}$/ $31.8_{ 2.7}$ & $18.9_{ 0.6}$/ $78.3_{ 4.4}$/ $30.5_{ 0.9}$ & $19.6_{ 1.3}$/ $68.8_{ 3.9}$/ $30.5_{ 1.6}$ & $17.3_{ 1.4}$/ $75.4_{ 1.2}$/ $28.2_{ 1.9}$ & $25.5_{ 2.2}$/ $77.9_{ 3.5}$/ \textbf{38.4}$_{ 2.6}$ & $ 9.6_{ 1.1}$/ $ 48.6_{ 1.9}$/ $ 16.1_{ 1.6}$\\
\textbf{ChemProt} & $17.9_{ 2.2}$/ $62.0_{ 3.9}$/ $27.7_{ 2.9}$ & $18.7_{ 0.9}$/ $59.4_{ 5.0}$/ $28.4_{ 0.9}$ & $18.1_{ 0.7}$/ $56.8_{ 1.6}$/ $27.5_{ 0.7}$ & $19.0_{ 6.8}$/ $60.6_{ 8.2}$/ $28.7_{ 8.7}$ & $22.0_{ 0.3}$/ $69.7_{ 1.2}$/ \textbf{33.4}$_{ 0.4}$ & $ 15.9_{ 0.8}$/ $ 68.9_{ 1.9}$/ $ 25.9_{ 1.3}$\\
\textbf{GAD} & $63.7_{ 6.6}$/ $57.2_{ 7.9}$/ $60.2_{ 7.4}$ & $60.5_{ 5.0}$/ $62.8_{ 14.3}$/ $61.2_{ 8.1}$ & $60.2_{ 1.4}$/ $71.2_{ 20.1}$/ $64.4_{ 9.1}$ & $63.2_{ 5.8}$/ $72.7_{ 5.7}$/ $67.6_{ 5.8}$ & $64.1_{ 4.0}$/ $78.5_{ 11.5}$/ \textbf{70.3}$_{ 5.6}$ & $ 51.4_{ 0.9}$/ $ 92.3_{ 4.2}$/ $ 66.0_{ 1.8}$\\\midrule
\textbf{RE Average} & $33.8_{ 2.0}$/ $66.1_{ 2.8}$/ $39.9_{ 2.2}$ & $32.7_{ 1.7}$/ $66.8_{ 5.1}$/ $40.0_{ 2.7}$ & $35.1_{ 4.6}$/ $68.0_{ 10.7}$/ $43.5_{ 7.7}$ & $33.2_{ 0.6}$/ $69.6_{ 2.3}$/ $41.5_{ 1.4}$ & $37.2_{ 1.8}$/ $75.4_{ 4.5}$/ \textbf{47.4}$_{ 1.9}$ & $ 25.6_{ 0.1}$/ $ 70.0_{ 1.4}$/ $ 36.0_{ 0.4}$ \\ \bottomrule
\end{tabular}}
\caption{Main experimental results from Table \ref{tab:master} with additional results from BioBERT and RoBERTa base models for appropriate comparison.} 
\label{tab:master-base}
\end{table*}

\begin{figure*}[!th]
     \centering
     \begin{subfigure}[h]{0.4\textwidth}
         \centering
         \includegraphics[width=\columnwidth]{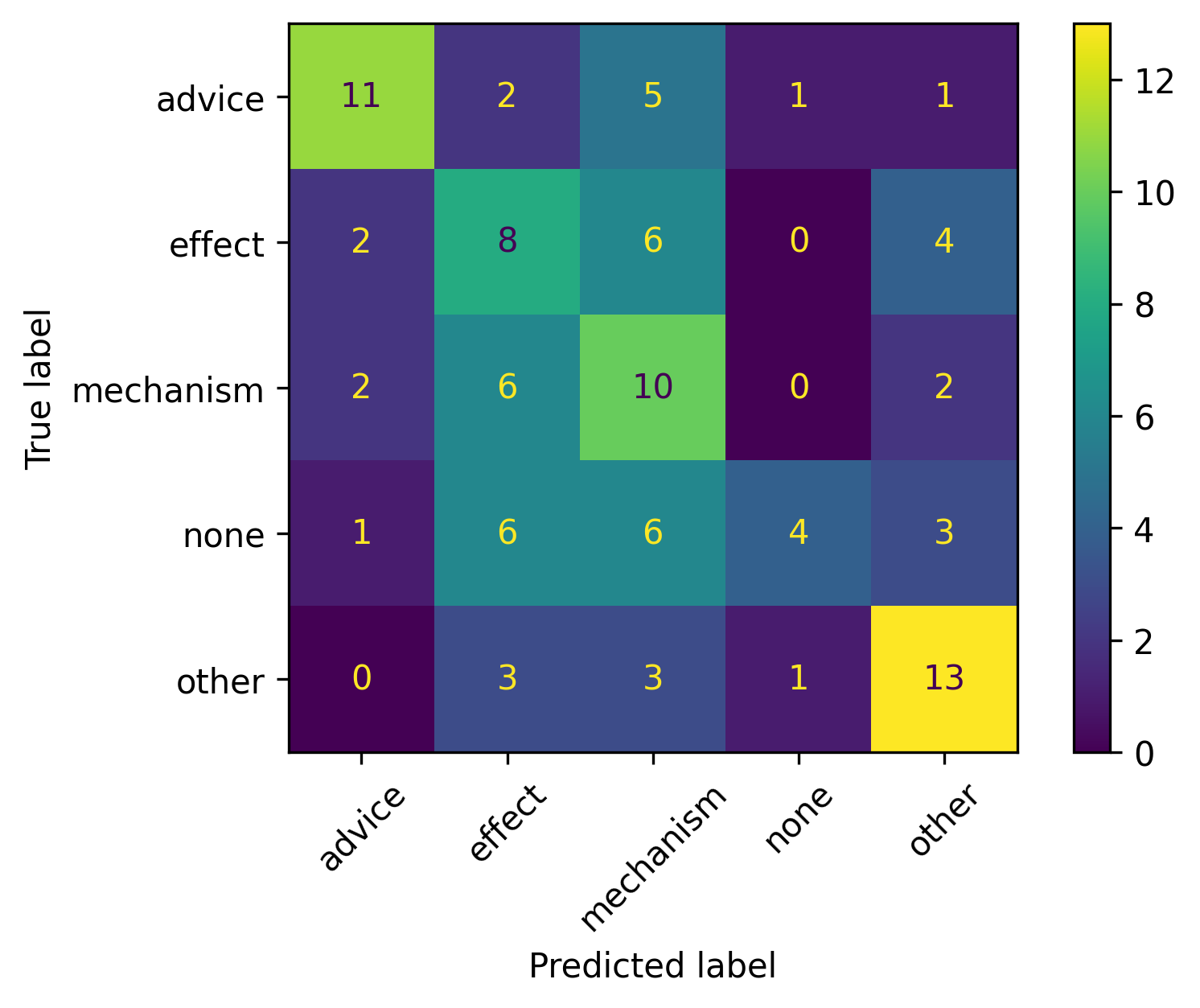}
         \label{fig:ddi_cm_gpt3}
     \end{subfigure}
     ~
     \begin{subfigure}[h]{0.4\textwidth}
         \centering
         \includegraphics[width=\columnwidth]{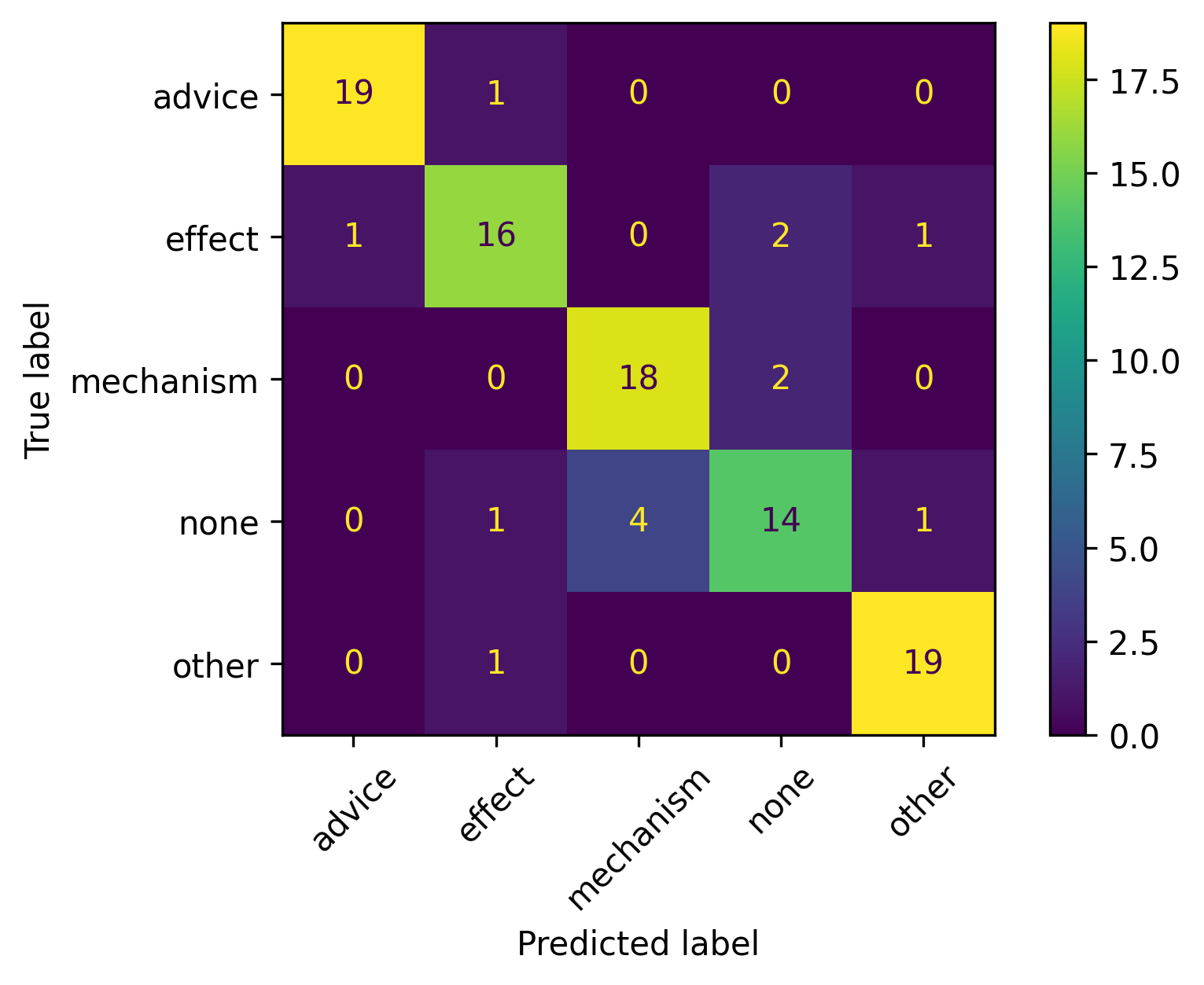}
         \label{fig:ddi_cm_roberta}
     \end{subfigure}
     \vspace{-15pt}
        \caption{Confusion matrices on \num{100} validation examples from DDI for GPT-3 (left) and RoBERTa-large (right).}
        \label{fig:DDI Confusion}
\vspace{-5pt}
\end{figure*}

\section{Base Models}\label{sec:appendix-base-models}
As expected, the base models added to Table \ref{tab:master-base} underperform their large counterparts on almost all datasets. Consistent with previous work \cite{Gu2022DomainSpecificLM}, benefiting from the biomedical-specific vocabulary, PubMedBERT-base handily outperforms other base models on the NER task (as well as some large models on several tasks). However, on the RE tasks, RoBERTa models perform the best. 
Since RE tasks requires more holistic understanding of the whole sentence, this suggests that RoBERTa provides more general linguistic knowledge than other PLMs specific to biomedicine.

\section{DDI Error Analysis}\label{sec:re_error_appendix}

\subsection{Confusion Matrices}\label{sec:re_confusion_analysis}

Figure \ref{fig:DDI Confusion} shows the error distribution for both GPT-3 and RoBERTa-large in a 100 example training set for the DDI relation extraction dataset. We obtain these by combining all folds from \num{5}-fold and leave-one-out cross-validation for RoBERTa-large and GPT-3 respectively. From the figure, we can see that GPT-3 in-context learning rarely predicts the \texttt{none} class which indicates two drugs bare no relation to each other. We note that RoBERTa-large also suffers from a larger error rate for the \texttt{none} class than other classes, indicating that this class is challenging for both models, however, the gap is much smaller for RoBERTa than GPT-3 in-context learning.

\begin{table*}[!t]
\small
\renewcommand{\arraystretch}{1.2}
\resizebox{\textwidth}{!}{
\begin{tabular}{@{}clcc@{}}
\toprule
\textbf{Label}             & \textbf{Example} & \textbf{Model} & \textbf{Correct}\\\midrule
\midrule
\multirow{3}{*}{Advice} &  \begin{tabular}[c]{@{}l@{}}\hlc[cyan!50!white]{Concomitant} \hlc[cyan!32!white]{use} \hlc[cyan!17!white]{of} \hlc[cyan!0!white]{\textbf{bromocriptine mesylate}} with other \hlc[cyan!0!white]{\textbf{ergot alkaloids}} \hlc[cyan!20!white]{is} \hlc[cyan!20!white]{not} \hlc[cyan!42!white]{recommended}. \end{tabular} & RoBERTa-large  & \cmark\\ \cmidrule(l){2-4} &  \begin{tabular}[c]{@{}l@{}}\hlc[cyan!28!white]{Concomitant} \hlc[cyan!5!white]{use} \hlc[cyan!7!white]{of} \hlc[cyan!0!white]{\textbf{bromocriptine}} \hlc[cyan!0!white]{\textbf{mesylate}} with other \hlc[cyan!0!white]{\textbf{ergot}} \hlc[cyan!0!white]{\textbf{alkaloids}} is not \hlc[cyan!50!white]{recommended}.\end{tabular} & GPT-3 & \cmark\\\midrule

\multirow{3}{*}{Advice} &  \begin{tabular}[c]{@{}l@{}}Consequently, it \hlc[cyan!45!white]{is} \hlc[cyan!32!white]{recommended} \hlc[cyan!50!white]{not} to exceed a single 2.\hlc[cyan!0!white]{5} mg \hlc[cyan!0!white]{\textbf{Vardenafil}} dose in a 72-hour period when used in combination with \\ \hlc[cyan!0!white]{\textbf{ritonavir}}. \end{tabular} & RoBERTa-large  & \cmark\\ \cmidrule(l){2-4} &  \begin{tabular}[c]{@{}l@{}}\hlc[cyan!19!white]{Consequently}, it \hlc[cyan!19!white]{is} \hlc[cyan!46!white]{recommended} \hlc[cyan!50!white]{not} \hlc[cyan!10!white]{to} exceed a \hlc[cyan!0!white]{single} 2.5 mg \hlc[cyan!0!white]{\textbf{Vardenafil}} dose \hlc[cyan!0!white]{in} a 72-hour period \hlc[cyan!0!white]{when} used \hlc[cyan!0!white]{in} combination with \\ \hlc[cyan!0!white]{\textbf{ritonavir}}.\end{tabular} & GPT-3 & \cmark\\\midrule

\multirow{3}{*}{Effect} &  \begin{tabular}[c]{@{}l@{}}However, reports suggest that \hlc[cyan!0!white]{\textbf{NSAIDs}} may diminish \hlc[cyan!41!white]{the} \hlc[cyan!45!white]{antihypertensive} \hlc[cyan!50!white]{effect} of \hlc[cyan!0!white]{\textbf{ACE inhibitors}}. \end{tabular} & RoBERTa-large  & \cmark\\ \cmidrule(l){2-4} &  \begin{tabular}[c]{@{}l@{}}\hlc[cyan!50!white]{However}, \hlc[cyan!29!white]{reports} \hlc[cyan!21!white]{suggest} \hlc[cyan!23!white]{that} \hlc[cyan!0!white]{\textbf{NSAIDs}} \hlc[cyan!7!white]{may} diminish the \hlc[cyan!15!white]{antihypertensive} \hlc[cyan!18!white]{effect} \hlc[cyan!13!white]{of} \hlc[cyan!0!white]{\textbf{ACE}} \hlc[cyan!0!white]{\textbf{inhibitors}}.\end{tabular} & GPT-3 & \cmark\\\midrule

\multirow{3}{*}{None} &  \begin{tabular}[c]{@{}l@{}}Although \hlc[cyan!7!white]{specific} \hlc[cyan!31!white]{studies} \hlc[cyan!10!white]{have} not been \hlc[cyan!17!white]{performed}, \hlc[cyan!34!white]{coadministration} \hlc[cyan!14!white]{with} \hlc[cyan!15!white]{drugs} \hlc[cyan!29!white]{that} \hlc[cyan!20!white]{are} \hlc[cyan!1!white]{mainly} metabolized \hlc[cyan!7!white]{by} \hlc[cyan!50!white]{CYP3A4} \\ (\hlc[cyan!8!white]{eg}, calcium channel blockers, dapsone, \hlc[cyan!0!white]{\textbf{disopyramide}}, quinine, amiodarone, quinidine, warfarin, \hlc[cyan!0!white]{\textbf{tacrolimus}}, cyclosporine, ergot \\ derivatives, pimozide, carbamazepine, fentanyl, alfentanyl, alprazolam, and triazolam) \hlc[cyan!1!white]{may} \hlc[cyan!10!white]{have} \hlc[cyan!16!white]{elevated} \hlc[cyan!11!white]{plasma} \\ \hlc[cyan!3!white]{concentrations} \hlc[cyan!34!white]{when} \hlc[cyan!17!white]{coadministered} \hlc[cyan!14!white]{with} \hlc[cyan!45!white]{saquinavir}; \end{tabular} & RoBERTa-large  & \cmark\\ \cmidrule(l){2-4} &  \begin{tabular}[c]{@{}l@{}}\hlc[cyan!6!white]{Although} \hlc[cyan!22!white]{specific} \hlc[cyan!27!white]{studies} \hlc[cyan!11!white]{have} \hlc[cyan!50!white]{not} \hlc[cyan!26!white]{been} \hlc[cyan!22!white]{performed}, coadministration \hlc[cyan!34!white]{with} \hlc[cyan!2!white]{drugs} \hlc[cyan!16!white]{that} are \hlc[cyan!6!white]{mainly} \hlc[cyan!43!white]{metabolized} \hlc[cyan!17!white]{by} CYP3A4 \\ (eg, calcium \hlc[cyan!13!white]{channel} blockers, dapsone, \hlc[cyan!0!white]{\textbf{disopyramide}}, quinine, amiodarone, \hlc[cyan!14!white]{quinidine}, warfarin, \hlc[cyan!0!white]{\textbf{tacrolimus}}, \hlc[cyan!8!white]{cyclosporine}, ergot \\ derivatives, pimozide, \hlc[cyan!19!white]{carbamazepine}, fentanyl, alfentanyl, alprazolam, and \hlc[cyan!1!white]{triazolam}) may \hlc[cyan!11!white]{have} \hlc[cyan!33!white]{elevated} plasma \\ concentrations when \hlc[cyan!17!white]{coadministered} \hlc[cyan!34!white]{with} \hlc[cyan!39!white]{saquinavir};\end{tabular} & GPT-3 & \begin{tabular}[c]{@{}c@{}}\xmark\\(Other)\end{tabular}\\\midrule

\multirow{3}{*}{None} &  \begin{tabular}[c]{@{}l@{}}- \hlc[cyan!0!white]{Cholestyramine} and \hlc[cyan!2!white]{colestipol} \hlc[cyan!0!white]{resins}: \hlc[cyan!0!white]{\textbf{Cholestytamine}} and \hlc[cyan!2!white]{\textbf{colestipol}} \hlc[cyan!1!white]{\textbf{resins}} have \hlc[cyan!20!white]{the} \hlc[cyan!3!white]{potential} \hlc[cyan!13!white]{of} \hlc[cyan!22!white]{binding} \hlc[cyan!36!white]{thiazide} \hlc[cyan!26!white]{diuretics} and \\ \hlc[cyan!26!white]{reducing} \hlc[cyan!34!white]{diuretic} \hlc[cyan!21!white]{absorption} \hlc[cyan!15!white]{from} \hlc[cyan!20!white]{the} \hlc[cyan!50!white]{gastrointestinal} \hlc[cyan!0!white]{tract} \end{tabular} & RoBERTa-large  & \cmark\\ \cmidrule(l){2-4} &  \begin{tabular}[c]{@{}l@{}}- \hlc[cyan!0!white]{Cholestyramine} and \hlc[cyan!18!white]{colestipol} \hlc[cyan!0!white]{resins}: \textbf{Cholestytamine} and \hlc[cyan!0!white]{\textbf{colestipol}} \hlc[cyan!0!white]{\textbf{resins}} \hlc[cyan!7!white]{have} the potential of \hlc[cyan!46!white]{binding} \hlc[cyan!16!white]{thiazide} diuretics and \\ reducing \hlc[cyan!43!white]{diuretic} absorption \hlc[cyan!3!white]{from} the \hlc[cyan!50!white]{gastrointestinal} \hlc[cyan!39!white]{tract}\end{tabular} & GPT-3 & \begin{tabular}[c]{@{}c@{}}\xmark\\(Mechanism)\end{tabular}\\\midrule

\multirow{3}{*}{None} &  \begin{tabular}[c]{@{}l@{}}\hlc[cyan!28!white]{Monoamine} \hlc[cyan!27!white]{oxidase} (\hlc[cyan!1!white]{MAO}) inhibitors \hlc[cyan!6!white]{such} \hlc[cyan!0!white]{as} \hlc[cyan!0!white]{\textbf{isocarboxazid}} (e.g., Marplan), phenelzine (e.g., Nardil), procarbazine (e.g., \\ Matulane), selegiline (e.g., Eldepryl), and tranylcypromine (e.g., \textbf{Parnate}): Using \hlc[cyan!1!white]{these} \hlc[cyan!14!white]{medicines} \hlc[cyan!26!white]{with} \hlc[cyan!50!white]{L-tryptophan} \hlc[cyan!16!white]{may} \\ increase \hlc[cyan!0!white]{the} \hlc[cyan!11!white]{chance} \hlc[cyan!37!white]{of} \hlc[cyan!36!white]{side} \hlc[cyan!16!white]{effects}. \end{tabular} & RoBERTa-large  & \cmark\\ \cmidrule(l){2-4} &  \begin{tabular}[c]{@{}l@{}}Monoamine oxidase (MAO) \hlc[cyan!45!white]{inhibitors} \hlc[cyan!6!white]{such} \hlc[cyan!21!white]{as} \hlc[cyan!0!white]{\textbf{isocarboxazid}} (e.g., Marplan), phenelzine (e.g., \hlc[cyan!2!white]{Nardil}), procarbazine (e.g., \\ Matulane), selegiline (e.g., Eldepryl), \hlc[cyan!3!white]{and} tranylcypromine (e.g., \hlc[cyan!0!white]{\textbf{Parnate}}): \hlc[cyan!50!white]{Using} \hlc[cyan!12!white]{these} medicines \hlc[cyan!5!white]{with} \hlc[cyan!4!white]{L}-\hlc[cyan!8!white]{tryptophan} \hlc[cyan!17!white]{may} \\ \hlc[cyan!40!white]{increase} \hlc[cyan!35!white]{the} \hlc[cyan!20!white]{chance} of \hlc[cyan!36!white]{side} \hlc[cyan!0!white]{effects}.\end{tabular} & GPT-3 & \begin{tabular}[c]{@{}c@{}}\xmark\\(Effect)\end{tabular}\\

\bottomrule
\end{tabular}}
\caption{LIME-based saliency scores for more DDI examples. We present 3 examples with true drug-drug interactions predicted correctly by both models and 3 \texttt{none} examples predicted correctly by RoBERTa-large but incorrectly by GPT-3 in-context learning. As in Table \ref{tbl:re_error_analysis}, masking out words highlighted in \textcolor{cyan}{blue} changes the model's current prediction and the color's intensity indicates the strength of the effect on the final prediction. The drugs shown in \textbf{bold} are the head and tail entities for the relation being queried.}
\label{tbl:more_re_error_examples}
\vspace{-5pt}
\end{table*}

\subsection{Qualitative Analysis}\label{sec:qualitative}
In Table \ref{tbl:more_re_error_examples}, we present 3 positive and 3 \texttt{none} DDI examples respectively to help illustrate the more challenging nature of the \texttt{none} class as well as RoBERTa-large's superior ability to attend to more relevant implicit signals. In all three positive examples, the saliency scores attributed by LIME for RoBERTa and GPT-3 agree closely, suggesting that both models are able to leverage relevant surface level signals. The feature attribution for the \texttt{none} examples, however, suggests that GPT-3 continues leveraging surface level signals when more complex sentence level information is needed which RoBERTa seems to extract and use more effectively.

The first \texttt{none} example shows GPT-3's prediction is affected by several irrelevant features such as other drugs in the drug list (``\textit{channel}'', ``\textit{quinidine}'' and ``\textit{carbamazepine}''), the initial phrase explaining that specific studies have not been performed and the word ``\textit{metabolized}''. In contrast, RoBERTa is unaffected by the removal of drugs from the drug list and is correctly affected by important signals such as the removal of ``\textit{CYP3A4 (eg.}'', similar to the example in Table \ref{tbl:re_error_analysis}.
For the second \texttt{none} example, GPT-3's incorrect prediction is most strongly affected by the words ``\textit{binding}'', ``\textit{diuretic}'' and ``\textit{gastrointestinal}'' while for RoBERTa the effect of removing words is more uniformly distributed over the phrase ``\textit{binding thiazide diuretics and reducing diuretic absorption from the gastrointenstinal tract}''. This indicates that RoBERTa's prediction relies on broader phrase level information rather than word level signals. 
In the last example, we note that removing the phrase ``\textit{with L-tryptophan}'' from the sentence would create an interaction between the drugs being queried by yielding the phrase ``\textit{Using these medicines may increase the chance of side effects}''. The fact that RoBERTa's prediction is strongly affected by the removal of this phrase indicates that its decision boundary uses more complex linguistic signals than GPT-3 which leverages single words such as ``\textit{inhibitors}'', ``\textit{Using}'' and ``\textit{increase}'' to arrive at its prediction.

\subsection{LIME Details}\label{sec:lime}

We choose LIME \cite{lime} to perform our RE error analysis because it enables us to obtain faithful local explanations for GPT-3 in-context learning which are directly comparable with the ones from RoBERTa or other small PLMs. We use a modified version of the original LIME implementation\footnote{\url{https://github.com/marcotcr/lime}} \cite{lime} to carry out our analysis in  Appendix~\ref{sec:qualitative} and \S\ref{sec:re_error_analysis}. Due to resource constraints, we modify the token removal method in the original implementation from randomly masking out tokens to a sliding window of 3 tokens. This allows us to look at how phrase removal changes predictions while still using a reasonable number of neighbor examples. Since we use this tool for analyzing relation extraction only, we do not remove the entities that are being queried. For GPT-3 in-context learning, we keep the few-shot prompts constant and use \texttt{BLANK} as the replacement token given that GPT-3 does not have a mask token. We do not observe a large difference in the saliency scores when this mask token was changed. In our visualizations, the saliency score for each word is normalized by the largest score found for that example in order to make effects more apparent. 

\end{document}